\documentclass[a4paper,fleqn]{cas-dc}

\usepackage[numbers]{natbib}
\usepackage{subfigure}
\usepackage{caption}
\usepackage{tabularx}
\def\tsc#1{\csdef{#1}{\textsc{\lowercase{#1}}\xspace}}
\tsc{WGM}
\tsc{QE}
\tsc{EP}
\tsc{PMS}
\tsc{BEC}
\tsc{DE}

\begin{document}

\let\WriteBookmarks\relax
\def\floatpagepagefraction{1}
\def\textpagefraction{.001}
\shorttitle{A Dataset and Method for HVA Estimation}
\shortauthors{Ningyuan Xu et~al.}

\title [mode = title]{A Dataset and Method for Hallux Valgus Angle Estimation Based on Deep Learing}
%


\author[1,2]{Ningyuan Xu}[type=author,
                        orcid=0000-0002-5977-0187]
\fnmark[1]
\ead{xuningyuan@nimte.ac.cn}

\credit{Conceptualization of this study, Methodology, Software}

\address[1]{University of Chinese Academy of Sciences,
    No.19 Yuquan Road, Shijingshan District, Beijing, China}

\author[2]{Jiayan Zhuang}[
        orcid=0000-0002-8350-6116,
        ]
\cormark[1]
\fnmark[1]
\ead{zhuangjiayan@nimte.ac.cn}

\author[3,4]{Yaojun Wu}[
   ]
\ead{wyj3247@163.com}
\credit{Data curation, Writing - Original draft preparation}

\address[2]{Ningbo Institute of Industrial Technology, Chinese Academy of Sciences,
    No.1219, Zhongguan West Road, Zhenhai District, Ningbo City, Zhejiang Province, China}

\author[2]{Jiangjian Xiao}[
]

\ead{xiaojiangjian@nimte.ac.cn}

\address[3]{Hwa Mei Hospital, University of Chinese Academy of Sciences,
    41 Northwest Street, Ningbo City, Zhejiang Province, China}
\address[4]{Ningbo Institute of Life and health Industry, University of Chinese Academy of sciences}

\cortext[cor1]{Corresponding author}

\begin{abstract}
Angular measurements is essential to make a resonable treatment for Hallux valgus (HV), a common forefoot deformity. However, it still depends on manual labeling and measurement, which is time-consuming and sometimes unreliable. Automating this process is a thing of concern.\\
However, it lack of dataset and the keypoints based method which made a great success in pose estimation is not suitable for this field.
To solve the problems, we made a dataset and developed an algorithm based on deep learning and linear regression. It shows great fitting ability to the ground truth.
\end{abstract}

%


\maketitle

\section{Introduction}

Hallux valgus (HV) is a common forefoot deformity characterized by
a valgus deviation of the great toe and varus deviation of the first metatarsal\cite{AlvarezR1984}.
Its prevalence increases with age, and it afflicts 3.5\%
of adolescents, 23\% of adults aged 18-65 years and 35.7\%
of adults aged 65 years and older\cite{NixS2010,SpahnG2004}.
Patients with HV typically complain of over forefoot pain,
intolerance of shoe wear, impaired gait patterns or falls among the aged\cite{BenvenutiF1995,MenzHB2005,MenzHB2001,KoskiK1996}.
Up to date, there have been more than 100 different operative techniques defined for hallux valgus\cite{WagnerE2016}.
The proper selection of various operative procedures is commonly based on X-ray
angular measurement of the foot before operation, including the hallux valgus angle (HVA), the inter metatarsal angle (IMA),
and the distal metatarsal articular angle (DMAA)\cite{HeinemanN2020}. According to the range of HVA and IMA,
the severity of HV can be classified into three types\cite{Pique-VidalCarlos2019}:
\begin{enumerate}
    \item mild(15° $\leq$ HVA $\leq$ 20°, 9° $\leq$ IMA $\leq$ 11°)
    \item moderate (21° $\leq$ HVA $\leq$ 39°, 12° $\leq$ IMA $\leq$ 17°)
    \item severe (HVA $\geq$ 40°, IMA $\geq$ 18°)
\end{enumerate}
\begin{figure}
    \centering
    \subfigure[template]{
        \includegraphics[height=100pt]{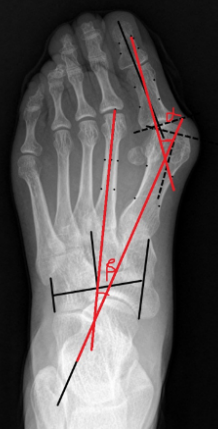}
    }
    \subfigure[samples]{
        \includegraphics[height=100pt]{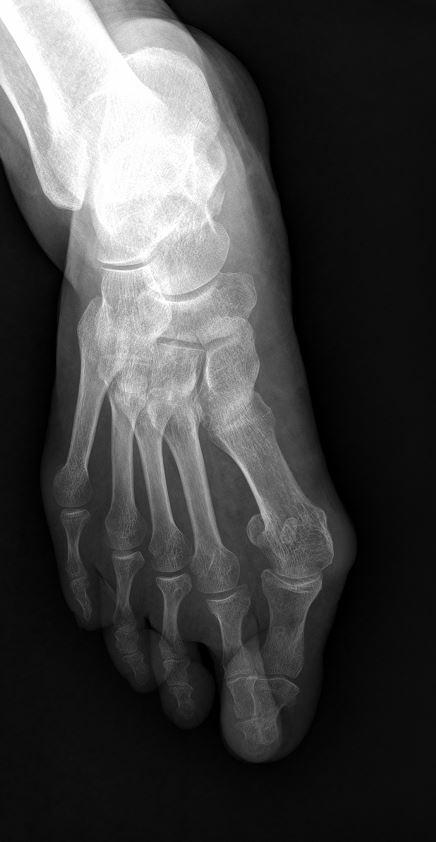}
        \includegraphics[height=100pt]{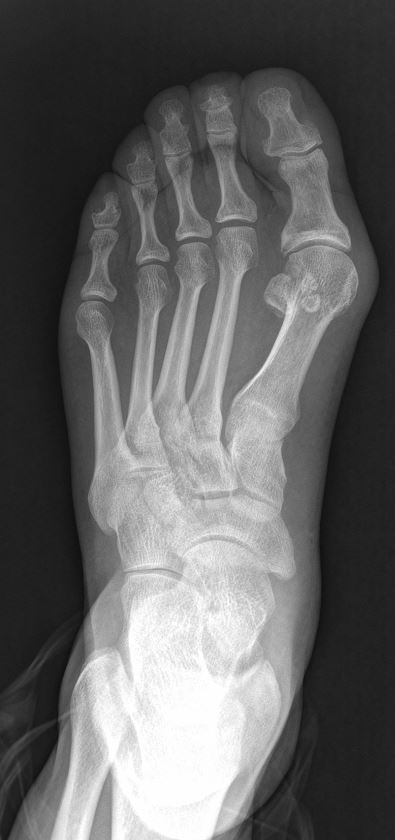}
        \includegraphics[height=100pt]{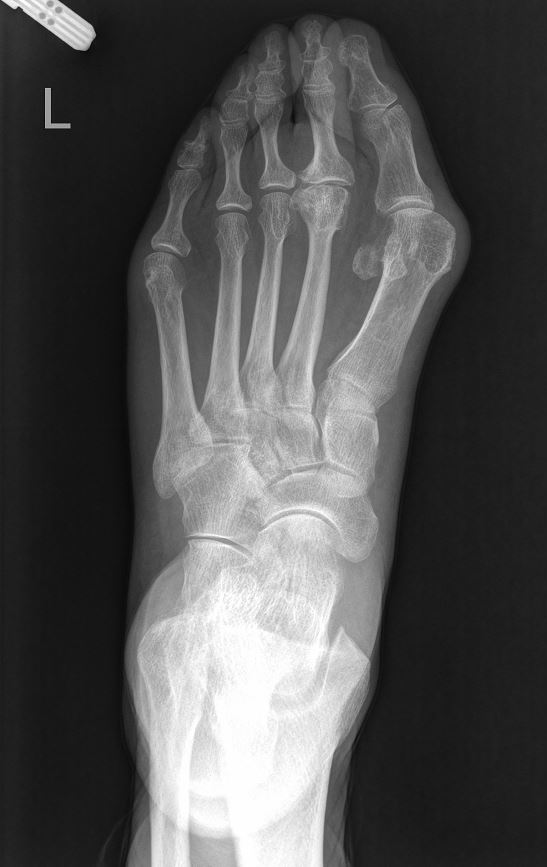}
    }
    \caption{Samples of Hallux valgus, (a) is a template labeled by doctor;
        (b) is several samples of our dataset}
    \label{fig:1.1}
\end{figure}
In order to make applicable treatment decisions, angular measurements must be accurate,
reliable, and reproducible. However, traditional X-ray measurements may be inaccurate
because they are influenced by variations in measurement techniques,
and technicians’ level of experience and ability to read X-ray images\cite{LeeKM2012}.
Numerous studies have showed intra observer and inter observer measurement errors\cite{LeeKM2012,CoughlinMJ2001,ChiTD2002,CruzEP2017,vanDerWoudeP2019}.
At present, the common practice of orthopedic surgeons is just like Fig \ref{fig:1.1} (a) shows:
\begin{enumerate}
    \item Draw the four auxiliary points of each phalanx.
    \item Draw the center lines of each phalanx.
    \item Measure the angles of the center lines.
\end{enumerate}
The whole process stays in the manual stage, which is very complicated.
At present, deep learning has made breakthroughs in many related fields,
such as human posture recognition and gesture recognition.
But there is no articles about hallux valgus angle estimation(HVAE) with deep learning.
HVAD has difficulties as follows:
\begin{itemize}
    \item To date, there is no publicly available dataset for HVA estimation.
    \item The domain of x-ray is different with natural image.
    X-ray is grayscale image, while natural image is based on RGB.
    So the models pretrained on large open datasets like ImageNet
    and COCO can’t be well migrated to an X-Ray scene.
    \item The Method based on the key points is not suitable with the scenario.
    Once the key points are off by a few pixels, the angle of the whole line will be greatly offset.
    A deviation of a few pixels is acceptable for key-point detection but not for angle estimation.
\end{itemize}
our contribution is as follows:
\begin{itemize}
    \item We are the first to use artificial intelligence to replace the traditional manual measurement of HVA and IMA.
    \item we made a HV dataset which collect from 143 patients and contains 235 preoperative images.
    \item We proposed a novel method based on neural network and traditional geometry. Compared to those method based on key points, our method can get more accurate estimation of HVA and IMA.
\end{itemize}
\begin{table*}
    \caption{Advantages and Disadvantages of Regression Based Method and Heatmap Based Method}
    \begin{tabularx}{\textwidth}{XXX}
        \toprule
        Framwork & Advantage & Disadvantage\\
        \midrule
        Direct regression based & Quick and direct, trained with an end-end fashion. Easy to be extended to 3D scenarios. & Difficult to learn mapping. Hard to be appled to multi-person case.\\
        Heatmap-based & Easy to be visualized. Robust to complicated case. & Large memory consumption for getting high resolution heat map. Hard to be extended to 3D scenarios.\\
        \bottomrule
    \end{tabularx}
    \label{Tab1}
\end{table*}
\section{Related Works}

Similar to the problem studied in this paper are human pose estimation,
animal pose estimation, face landmark detection, hand pose estimation and so on.
Traditional algorithms to deal with those problems used manual feature extraction
and complex human model to obtain local representation and global pose structure.\cite{DantoneM2013,GkioxariG2013}
After deep learning brought great innovation to this field,
the method can be grouped into two classes: One is based on regression,the other is based on heatmap.
We will introduce the development of both method detaily in section 2.1 and section 2.2.
\subsection{Models based on regression}

The method based on regression attempts to learn the mapping from the image
to the joint coordinates through the end-to-end framework
and generally generates the joint coordinates directly.\cite{ToshevA2014}
However, predicting joint coordinates directly is very difficult with few constraints.
So a more powerful network was introduced and the model structure was improved.Carreira $et al.$ \cite{CarreiraJ2016}
proposed an iterative error feedback network based on GoogLenet,
which recursively processes the combination of input images and output results.
The final results is improved from the initial coarse prediction. Sun $et al.$ \cite{SunX2017}
proposed a structural perceptual regression method based on ResNet-50.
The bone-based representation achieves more stable results by introduce body
structure information than using joint position alone.

\subsection{Models based on heatmap}
\begin{figure}
    \centering
    \includegraphics[width=0.5\textwidth]{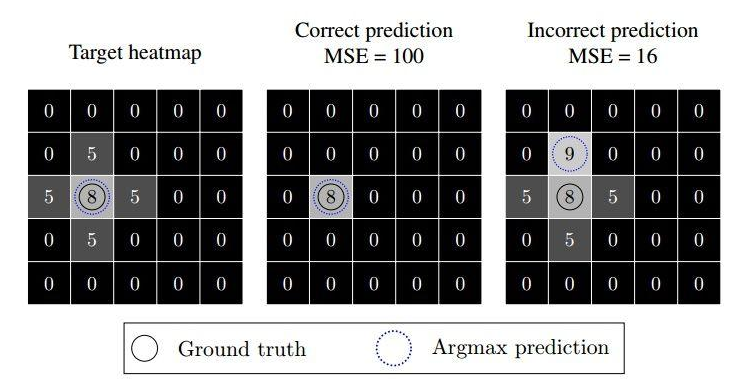}
    \caption{Inherent problems of Gaussian heatmap. The correct prediction have much bigger loss than incorrect prediction}
    \label{fig:2.1}
\end{figure}
Detection-based algorithms are dedicated to predicting the approximate position
of body parts or joints\cite{NewellA2016},
ground truth is usually a series of heatmaps(a two-dimensional Gaussian distribution centered on joint coordinates
to represent a joint position).\cite{WeiSE2016}
Although, obtaining joint coordinates from the heatmap is usually
a non-differentiable process, which hinders the end-to-end training of the network\cite{TompsonJ2014},
the heatmap representation is more robust than the coordinate representation,
most recent studies are based on the heatmap representation.\\
To get more accurate results, Papandreou $et al.$ \cite{PapandreouG2017} proposed
an improved joint position representation, which is a combination
of a binary activation heatmap and the corresponding position offset.
In order to make better use of the input information, the structure
of the neural network is very important. Some methods are mainly based on
classical networks with appropriate improvements, such as multi-scale
input network based on Googlenet \cite{RafiU2016} and deconvolution layer network
based on ResNet \cite{XiaoB2018}. \\
In terms of iterative refinement, Somework has designed a multi-stage network to refine rough prediction results
through end-to-end learning \cite{TompsonJ2015,BulatA2016,NewellA2016,WeiSE2016,YangW2017,BelagiannisV2017}.
Newell $et al.$ \cite{NewellA2016} proposed a layered Hourglass architecture
with residual modules as component units. Wei $et al.$ \cite{WeiSE2016} proposed a multi-stage
prediction framework with input images for each stage. Yang$et al.$ \cite{YangW2017}
designed a Pyramidal Residual Module (PRMS) to replace the Residual Module
of Hourglass Network, and enhanced the cross-scale invariance of DCNN by
learning features on different scales. Belagiannis and Zisserman\cite{BelagiannisV2017} combine
a 7-layer feedforward module with a recursive module to refine the results in an
iterative manner. The model learns to predict joint and limb position heatmaps.
The relationship between the visibility of key points and the real imbalance
of ground distribution is analyzed. In order to maintain a high resolution
representation of features across the entire network, Sun $et al.$ \cite{SunK2019} proposed
a new high resolution network with multi-scale feature fusion (HRNet).\\
As shown in \ref{Tab1}, both of two methods have their own advantages and disadvantages.
Direct regression learning for a single point is a highly nonlinear problem
and lacks robustness, while heatmap learning is supervised by
dense pixel information, so its robustness is better. However, compared with
the original image size, the resolution of the heatmap representation
is much lower due to the pooling operation in CNN, which limits the accuracy
of the joint coordinate estimation. What is worse, as Fig \ref{fig:2.1} shows, Gaussian heatmap can sometimes bias the optimization direction of the network, and get wrong results.
Luckily, our method can effectively avoid the influence of a few deviations on the final prediction results.

\section{Method}
\begin{figure*}
    \includegraphics[width=\textwidth]{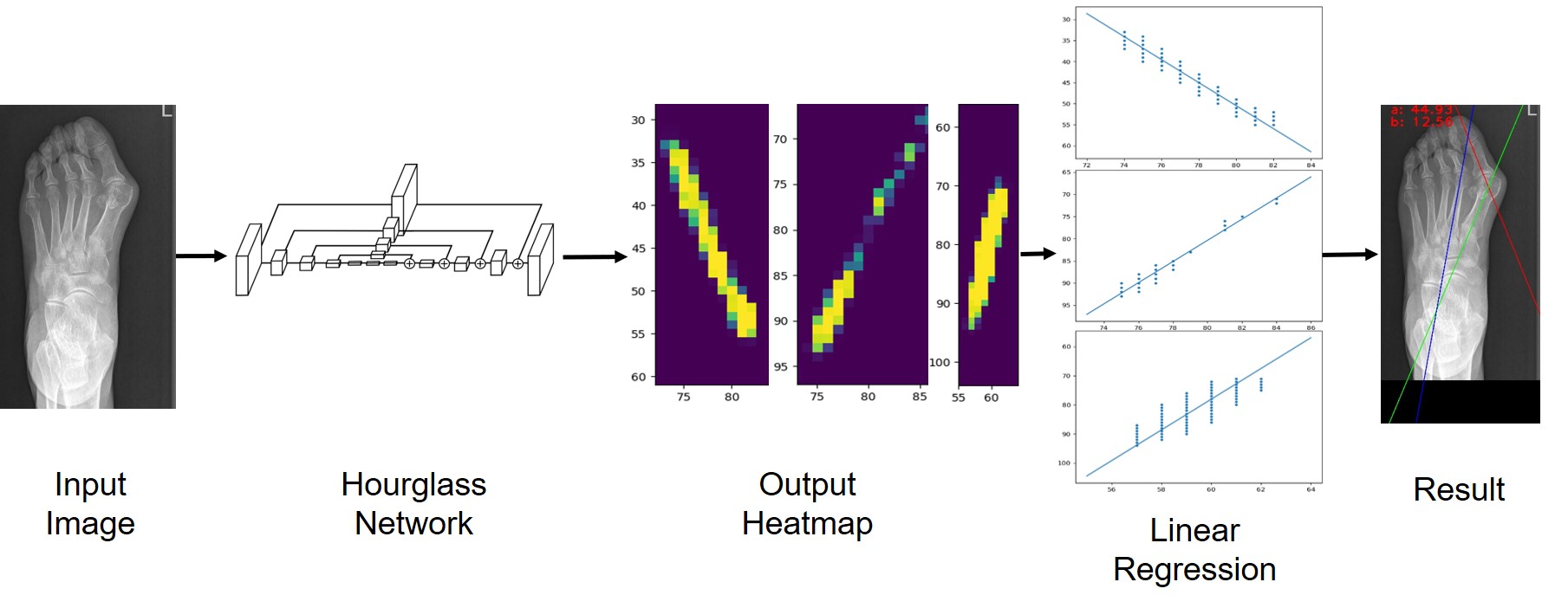}
    \caption{PipeLine of our method, our method is composed by neural network and linear regression.
    The neural network predicts three heatmaps from an input image,
        then the positions of points whose confidence is greater then 0.5 is picked up. The linear equations is calculated by linear regression on the points.
    With the linear equations, we can calculate the angles between the lines and visualize the result.}
    \label{fig:3.1}
\end{figure*}
The Pipeline of our method is as Fig \ref{fig:3.1} shows. It's based on deep learning and traditonal geometry.
In total, it can be divided into two parts: neural network and linear regression.
\subsection{Neural Network}
\begin{figure}
    \centering
    \subfigure[predictions]{
        \includegraphics[height=110pt]{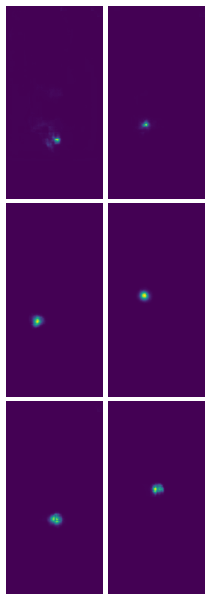}
        \includegraphics[height=110pt]{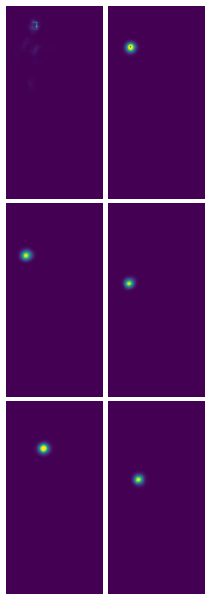}
    }
    \subfigure[results]{
        \includegraphics[height=110pt]{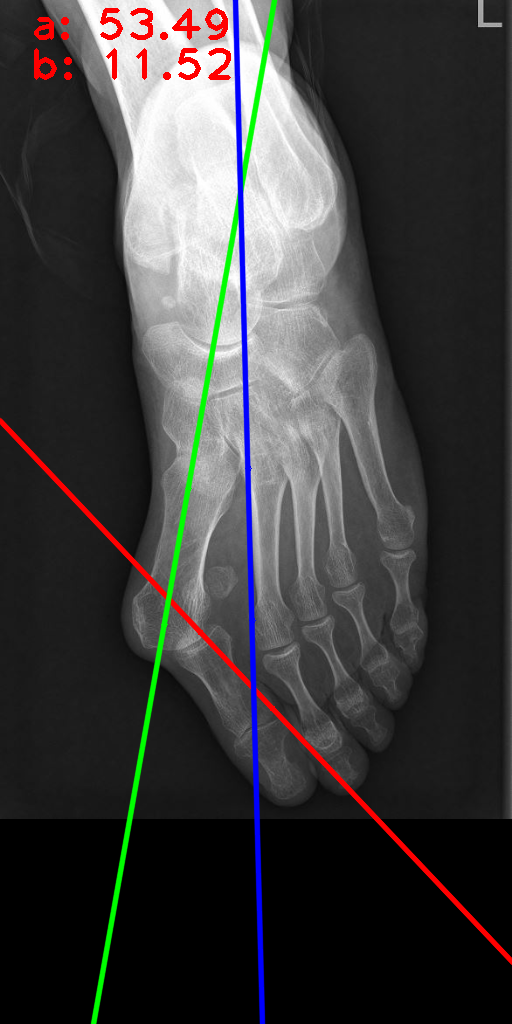}
        \includegraphics[height=110pt]{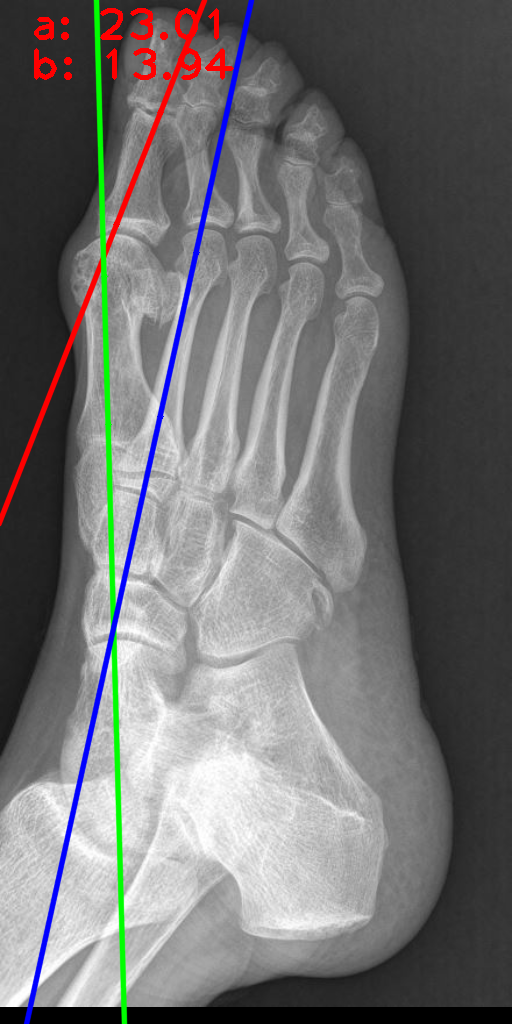}
    }
    \subfigure[labels]{
        \includegraphics[height=110pt]{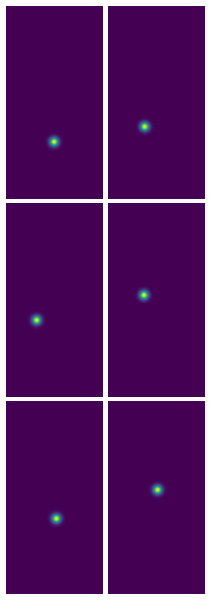}
        \includegraphics[height=110pt]{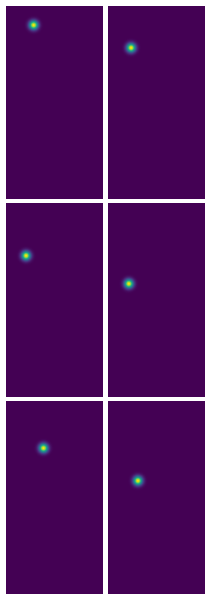}
    }
    \subfigure[ground truth]{
        \includegraphics[height=110pt]{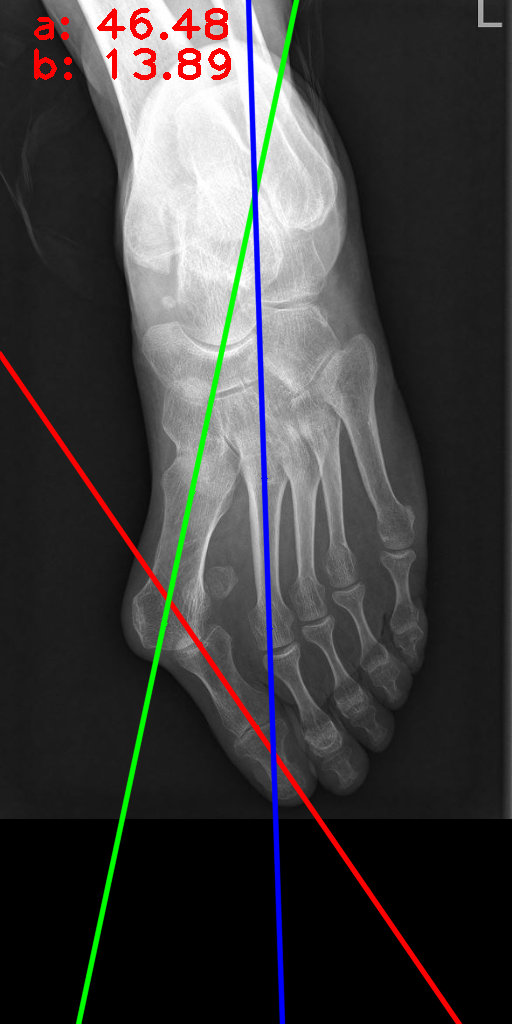}
        \includegraphics[height=110pt]{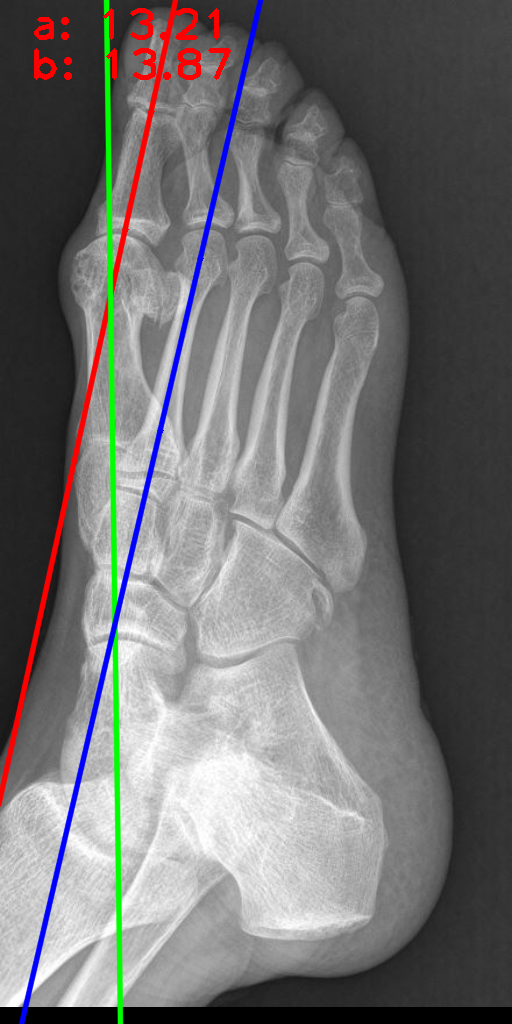}
    }
    \caption{The results of keypoints-based mothod. Fig (a) is the predicted heatmaps of keypoint detection model; (b) is the final results; (c) is the labels for keypoints detection model; (d) is the ground truth.  }
    \label{fig:3.2}
\end{figure}
\begin{figure}
    \centering
    \subfigure[2]{
        \includegraphics[width=0.1\textwidth]{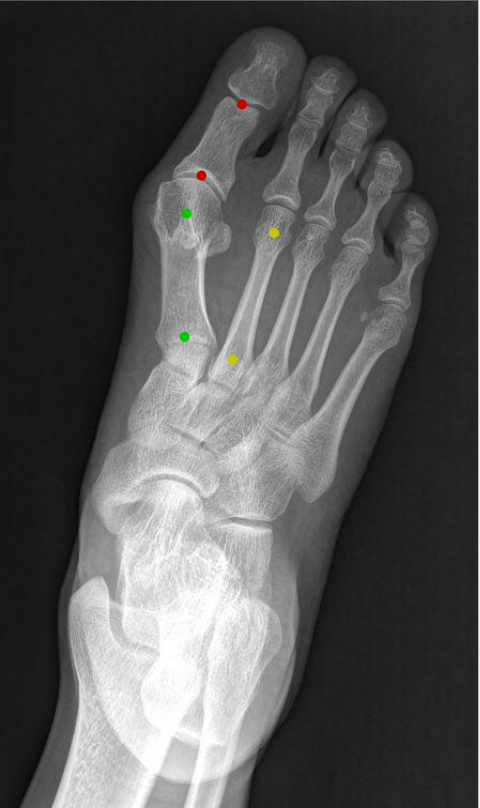}
    }
    \subfigure[3]{
        \includegraphics[width=0.1\textwidth]{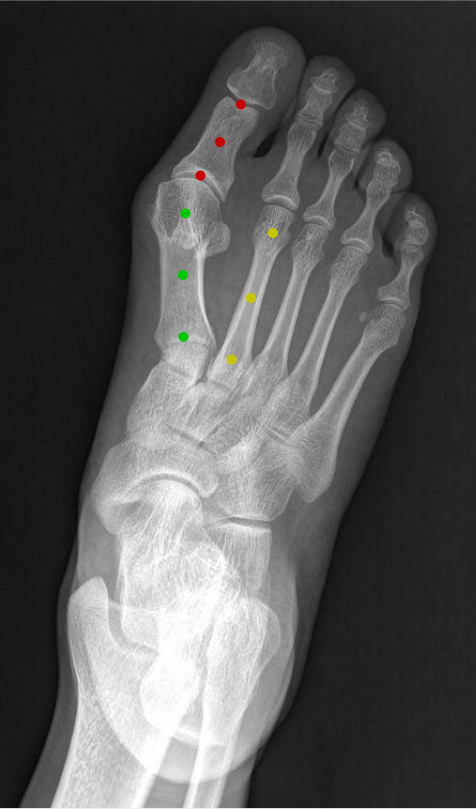}
    }
    \subfigure[4]{
        \includegraphics[width=0.1\textwidth]{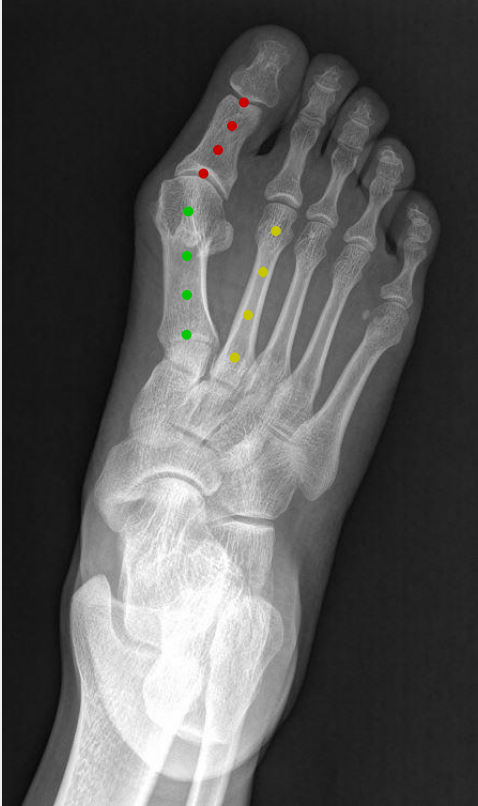}
    }
    \subfigure[infinity]{
        \includegraphics[width=0.1\textwidth]{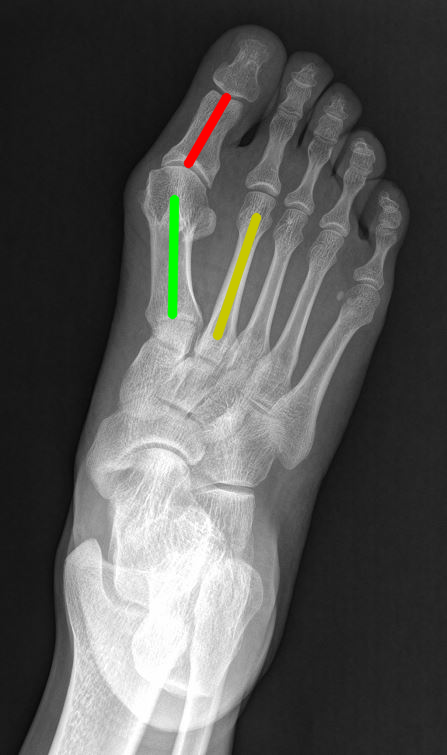}
    }
    \caption{Different Numbers of Keypoints of a Line Segment}
    \label{fig:3.3}
\end{figure}
\begin{figure}
\centering
\includegraphics[width=0.5\textwidth]{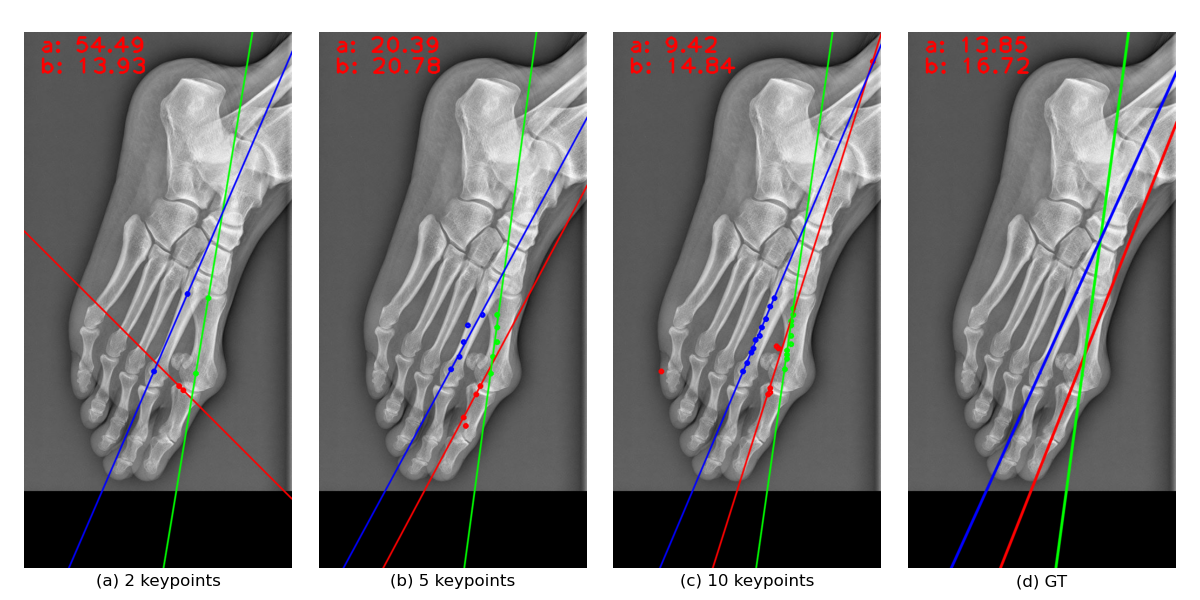}
\caption{Comparison of Results Based on Different Numbers of Keypoints }
\label{fig:3.4}
\end{figure}

\begin{figure}
    \centering
    \subfigure[predictions]{
        \includegraphics[height=140pt]{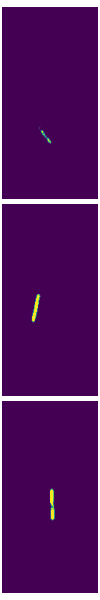}
        \includegraphics[height=140pt]{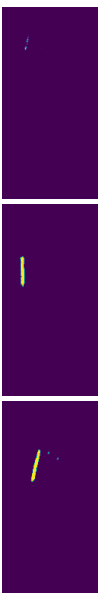}
    }
    \subfigure[results]{
        \includegraphics[height=140pt]{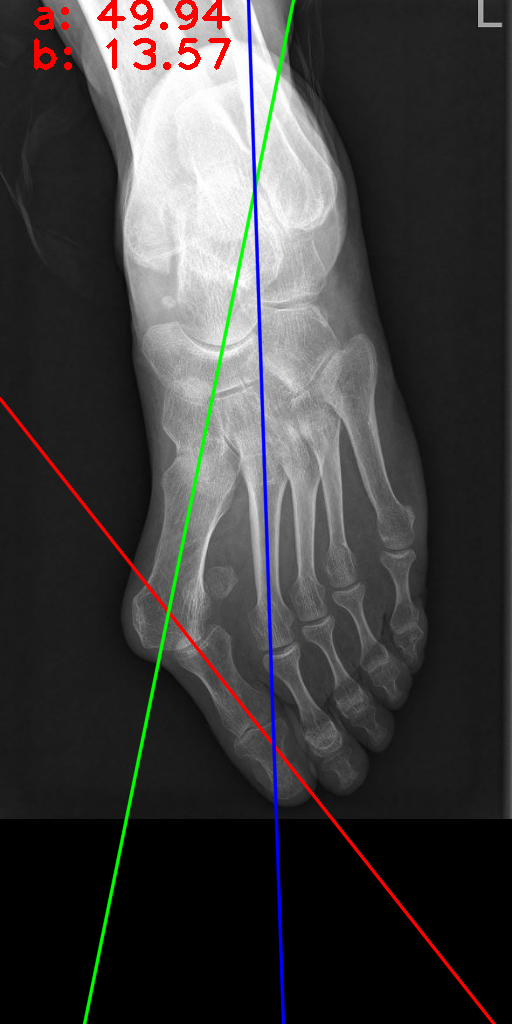}
        \includegraphics[height=140pt]{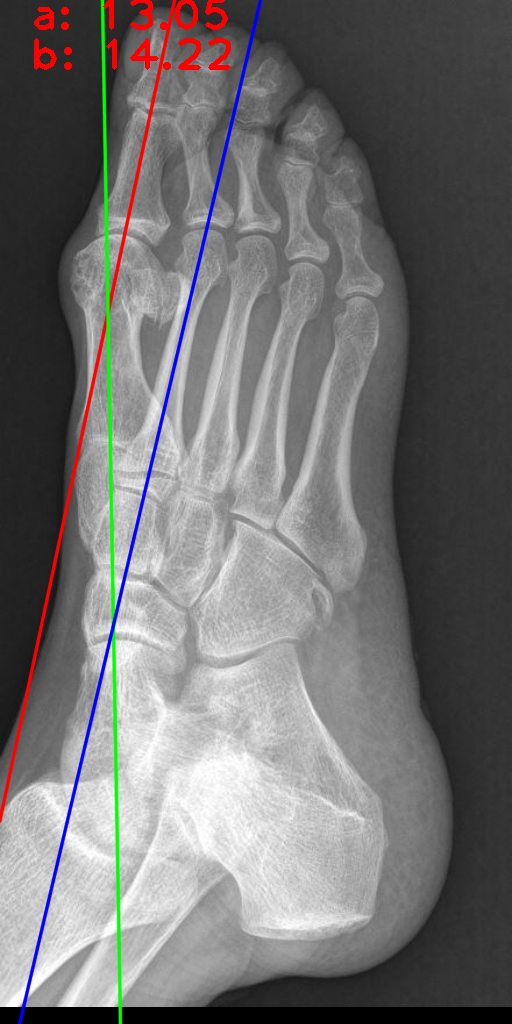}
    }
    \subfigure[labels]{
        \includegraphics[height=140pt]{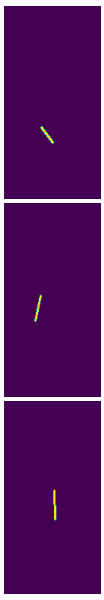}
        \includegraphics[height=140pt]{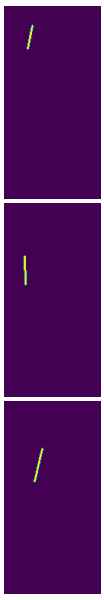}
    }
    \subfigure[ground truth]{
        \includegraphics[height=140pt]{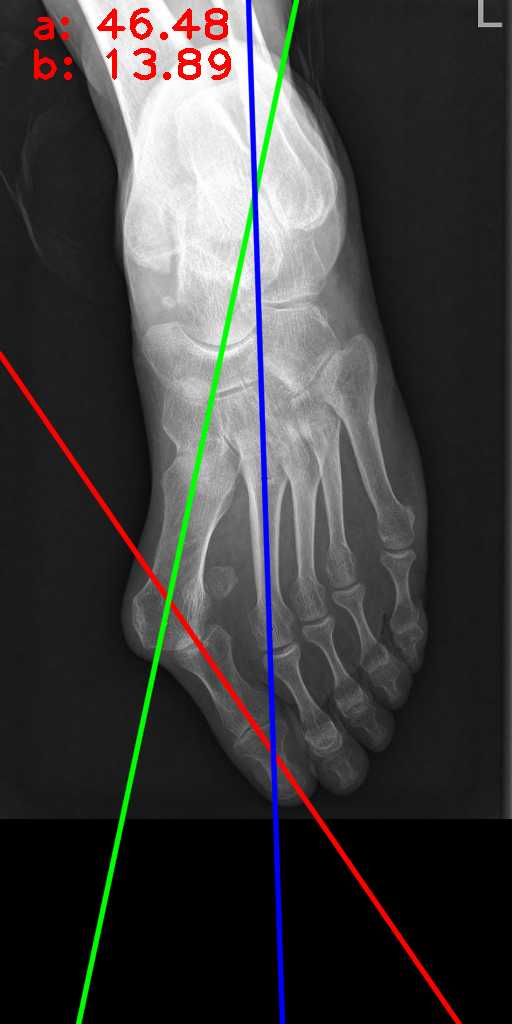}
        \includegraphics[height=140pt]{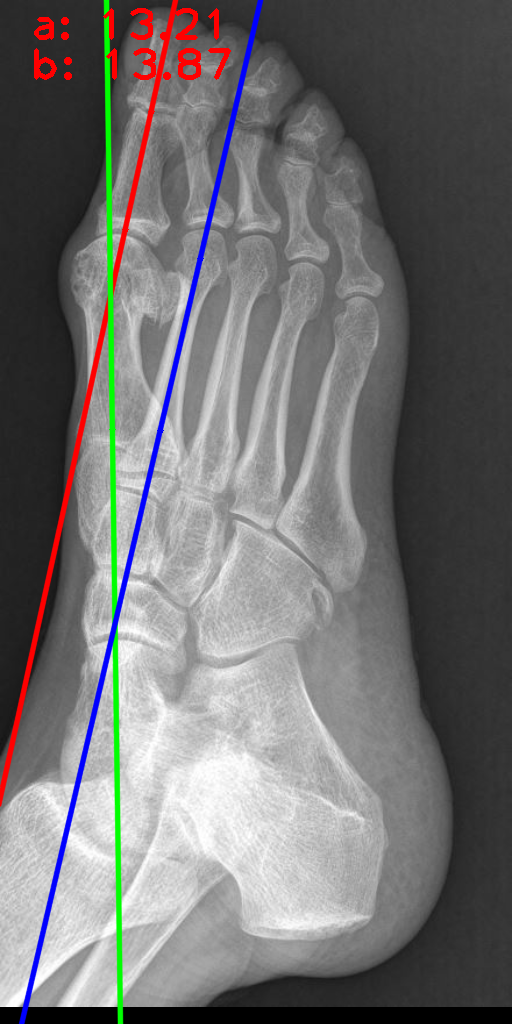}
    }
    \caption{The results of our method. Fig (a) is the predicted heatmaps of neural network; (b) is the final results after linear regression; (c) is the labels for neural network; (d) is the ground truth.}
    \label{fig:3.5}
\end{figure}
Our method adopts the classic  Hourglass neural network \cite{NewellA2016} in the field of key point
detection. Compared with other known backbones, such as Vgg16, Vgg19, ResNe,
DenseNet, the advantages of HG network lies in its symmetrical decoding
and encoding structure, which will output the heatmap zoom back to a quarter of
the original size, it can effectively improve the accuracy of predicting the keypoints.
When a line only depends on two keypoints, the pixel-level deviation
of keypoints can lead to a large deviation of the inclination of
the whole line, which make the deviation of the results intolerable. Even in the training dataset, the results are not ideal. As Fig \ref{fig:3.2} shows, the predicted heatmaps is closed with the labels in general position, but there is still pixel-level deviation, which lead the red lines on Fig \ref{fig:3.2} (b) deviated a lot compared to ground truth. The deviation of $\alpha$ reach around 7° and 10°.
To improve the situation, the first thing come to our mind is to take more keypoints. As Fig \ref{fig:3.3} shows, we can change numbers of the keypoints to 3, 4 and more. The results are also getting better as Fig \ref{fig:3.4} shows. But what if we take a limit on this case: filling in all the points between the two endpoints. It will certainly change into a line segment as Fig \ref{fig:3.3} (d) shows, and we don't need generate Guassian heatmaps for every points since the number is enough to mitigate the nonlinearity. We can simply place all the points of a line segment on one heatmap as Fig \ref{fig:3.5} (a) and (c) shows.\\
The problem of single Keypoint prediction is changed into the problem of front background segmentation.
Sigmoid activation function is used in the output layer to generate probabilities, the units whose probabilities greater than 0.5 are
the foreground (corresponding to the line segment), while the others is the background.
This method also makes the prediction result robust since the pre-background imbalance is alleviated.\\
More importantly, even if there are several points did not predict prospective, others correctly
predicted can help fit the ideal linear equation by linear regression. As \ref{fig:3.5} shows, although only a little points of red ones are detected by neural network, the final results after linear regression is still very good and the deviations of angles is much smaller.
\subsection{Linear Regression}
\begin{figure}
    \centering
    \subfigure[raw prediction]{
        \includegraphics[width=0.1\textwidth]{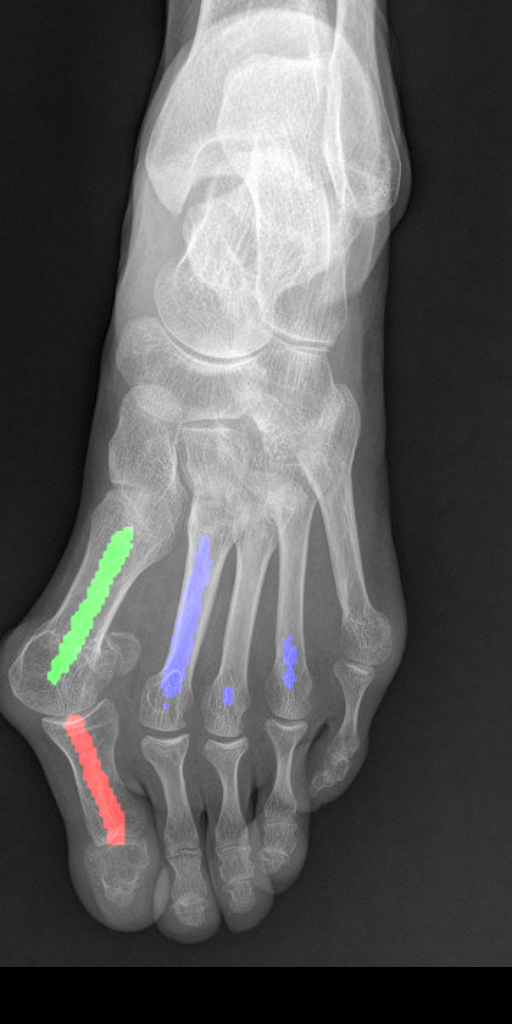}
    }
    \subfigure[results of L2]{
        \includegraphics[width=0.1\textwidth]{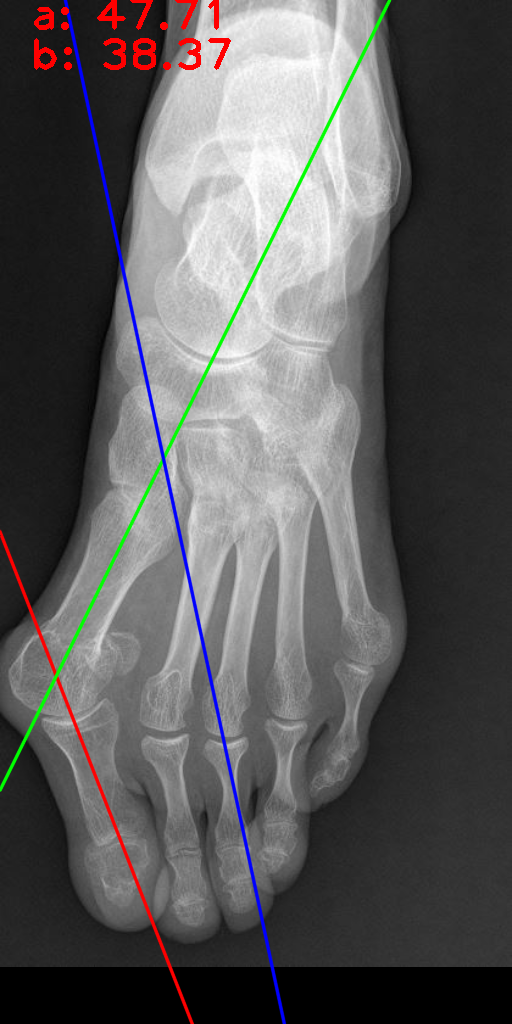}
    }
    \subfigure[results of WELSCH]{
        \includegraphics[width=0.1\textwidth]{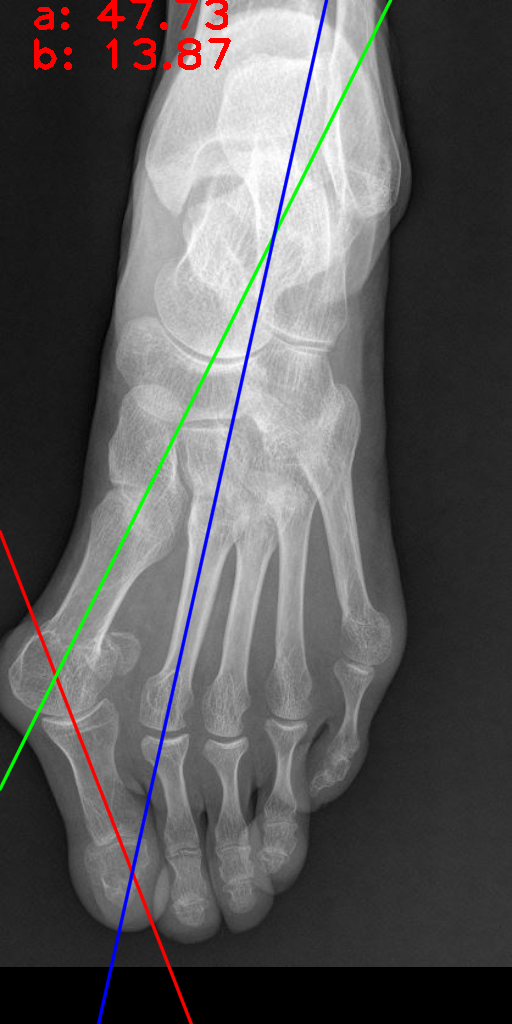}
    }
    \subfigure[GT]{
        \includegraphics[width=0.1\textwidth]{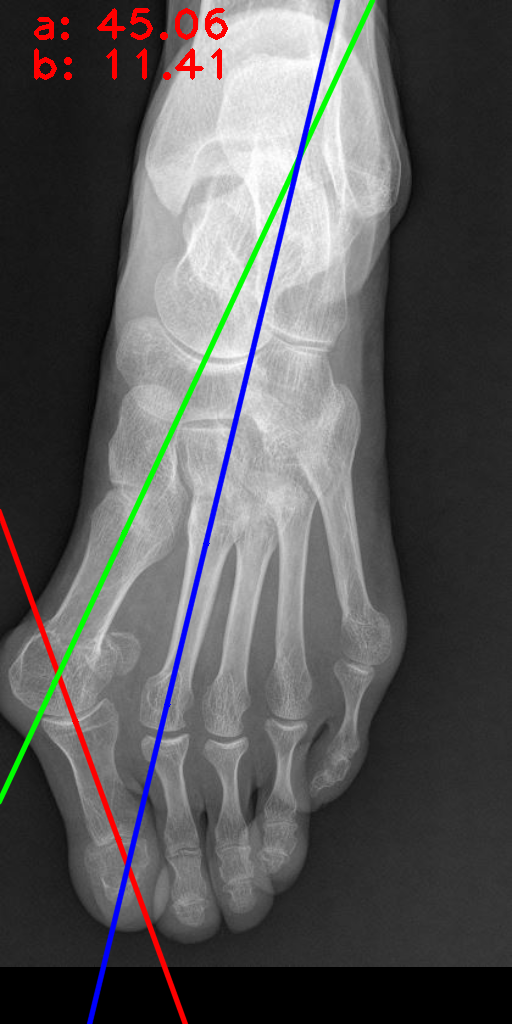}
    }
    \caption{Comparison of Regression results of Different Distance Metrics}
    \label{fig:3.6}
\end{figure}
In the part of linear regression, we first make a set of coordinates of units whose probabilities are greater than 0.5 through conditional judgment. Then we get the final linear equation by linear regression of the coordinate set. WELSCH loss instead of mean square loss(L2) is used to weaken the influence of several outliers. The commonly used distance metrics in linear regression are L2, L1, L12, FAIR, HUBER, WELSCH. L2 is too sensitive to outliers, L1, L12, FAIR and HUBER are linear or nearlt linear when the distance is far enough. Only WELSCH approaches a constant as the distance gets too far. Fig \ref{fig:3.6} is an example, (a) is the result of the network's predictions scaled and superimposed on the original image. The red ones and green ones are close to ground truth, but there are some outliers in the blue ones. When we take L2 as distance metric, the final result is as (b) shows, whose blue line is totally incorrect, when we take WELSCH insteaed, the results is much better as (c) shows.
\section{Experiments}
\subsection{Dataset}
We collect a HVA dataset which include 230 preoperative images from 143 patients. We labeled two endpoints for every phalanx that needs attention. We draw the center lines and calculated the angles $\alpha$ and $\beta$ for every image like \ref{fig:3.5} (d) shows. Here $a$ is $\alpha$, $b$ is $\beta$. Anyone who wants to make a better model can intuitively see the difference between their predictions and ground truth. In our experiments, we take 150 samples as training dataset, the rest 65 images as test dataset.
\subsection{Evaluation Metrics}
Because there are no ready-made evaluation metrics for our task, we developed a metric by ourselves. For this task, the most important goal is to get accurate and reliable angles. Mean angle errors(MAE) are first come to our mind, but there is an problem, it's very susceptible to outliers. For example, there are 10 samples and two method. For the first method, one of the results deviates by 90 degrees, the others deviates by 1 degree, then MAE becomes 9.9 degrees. For the second method, all of the results deviated by 9.9 degrees, so MAE is also 9.9 degrees. These two methods are indistinguishable under MAE. However, for surgeon, the first method have 9 acceptable results while the second has none. Therefore, we made a metric $acc^t$ which can present the accuracy of predicted angle errors less than some threshold.
The metric is as follows:
\begin{align}
    acc^t &= \frac{1}{total}\sum_{i=1}^{total}(P(X[i])-G(X[i])<t)
\end{align}
Where $t$ is an angle threshold, $total$ is the number of samples of dataset, $X$ is the images of dataset, $P(X[i])$ is the predicted angles, $G(X[i])$ is the manually annotated angles.
\subsection{Comparison of Different Line Width}
\begin{figure}
    \centering
        \subfigure[d=1,GT]{
            \includegraphics[width=0.04\textwidth]{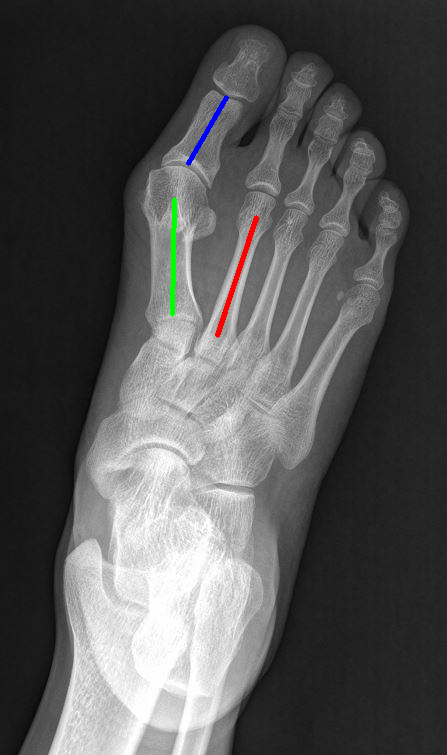}
        }
        \subfigure[d=1,c=1]{
            \includegraphics[width=0.04\textwidth]{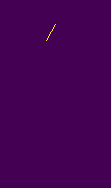}
        }
        \subfigure[d=1,c=2]{
            \includegraphics[width=0.04\textwidth]{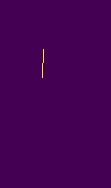}
        }
        \subfigure[d=1,c=3]{
            \includegraphics[width=0.04\textwidth]{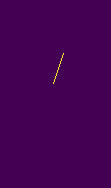}
        }
        \subfigure[d=2,GT]{
            \includegraphics[width=0.04\textwidth]{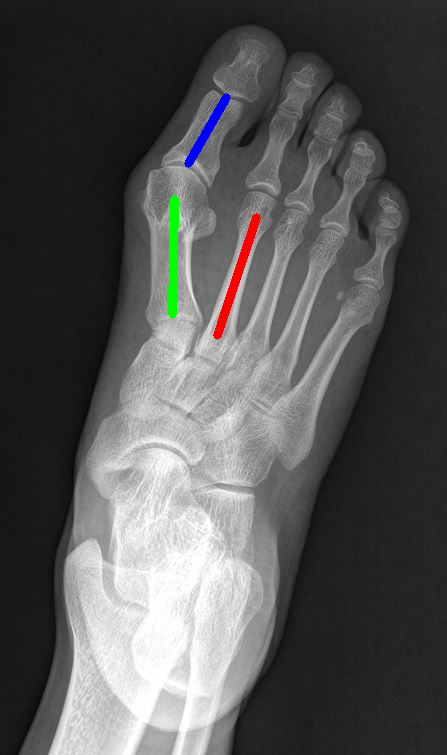}
        }
        \subfigure[d=2,c=1]{
            \includegraphics[width=0.04\textwidth]{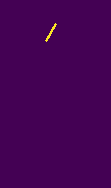}
        }
        \subfigure[d=2,c=2]{
            \includegraphics[width=0.04\textwidth]{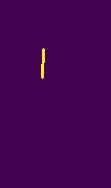}
        }
        \subfigure[d=2,c=3]{
            \includegraphics[width=0.04\textwidth]{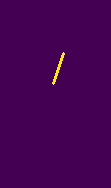}
        }
        \subfigure[d=4,GT]{
            \includegraphics[width=0.04\textwidth]{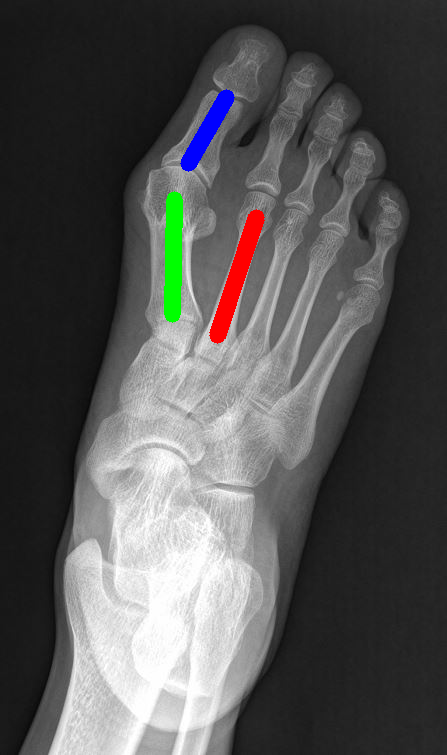}
        }
        \subfigure[d=4,c=1]{
            \includegraphics[width=0.04\textwidth]{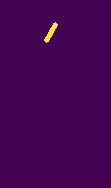}
        }
        \subfigure[d=4,c=2]{
            \includegraphics[width=0.04\textwidth]{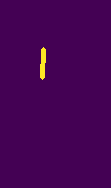}
        }
        \subfigure[d=4,c=3]{
            \includegraphics[width=0.04\textwidth]{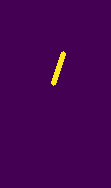}
        }
        \subfigure[d=8,GT]{
            \includegraphics[width=0.04\textwidth]{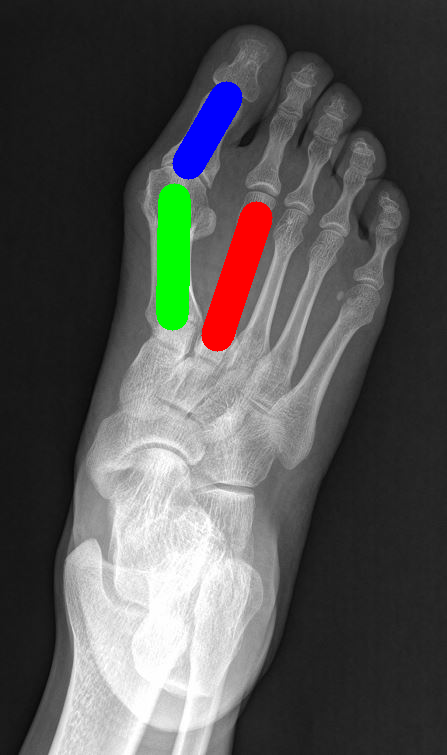}
        }
        \subfigure[d=8,c=1]{
            \includegraphics[width=0.04\textwidth]{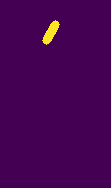}
        }
        \subfigure[d=8,c=2]{
            \includegraphics[width=0.04\textwidth]{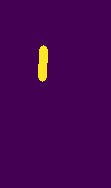}
        }
        \subfigure[d=8,c=3]{
            \includegraphics[width=0.04\textwidth]{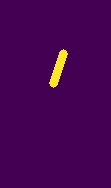}
        }
    \caption{Ground Truth and Labels of Different Line Width}
    \label{fig:4.1}
\end{figure}
\begin{table}
    \centering
    \caption{Comparison of Different Line Width on Test Dataset (Red is the best)}
    \includegraphics[width=0.5\textwidth]{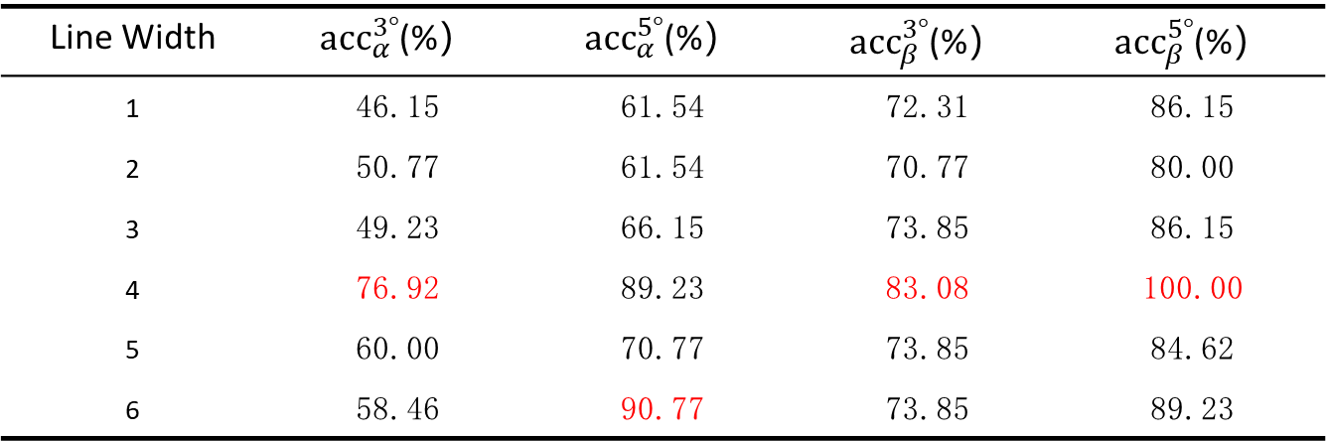}
    \label{tab:2}
\end{table}
We also take line width into consideration. A larger line width can make the trained network more robust, but at the same time it will also introduce more inherent deviations, especially for line segments with a smaller line length. The ground truth and labels of different line width are shown in \ref{fig:4.1}. Just looking at it with the naked eye, choosing a line width of 4 may be a good choice. To determine which line width is best, we did a comparative experiment. The results are shown in Tab \ref{tab:2}, $acc_{\alpha}^{3^\circ}$ means the accuracy of the prediction error of $\alpha <3^\circ $, the rest is similar. In general, it is indeed the best result when the line width is 4. From the Tab \ref{tab:2}, we can also find that the accuracy of $\beta$ is always higher than $\alpha$. It may because the phalanx corresponding to $\beta$ are more slender which makes them easier to predict.

\subsection{Results}

\begin{figure}
\centering
\subfigure[Result]{
    \includegraphics[width=0.1\textwidth]{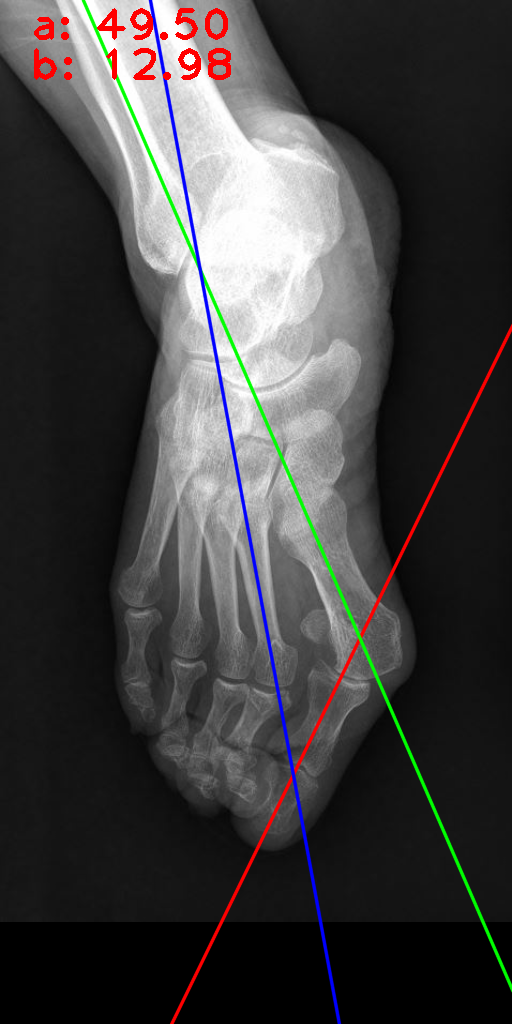}
}
\subfigure[GT]{
    \includegraphics[width=0.1\textwidth]{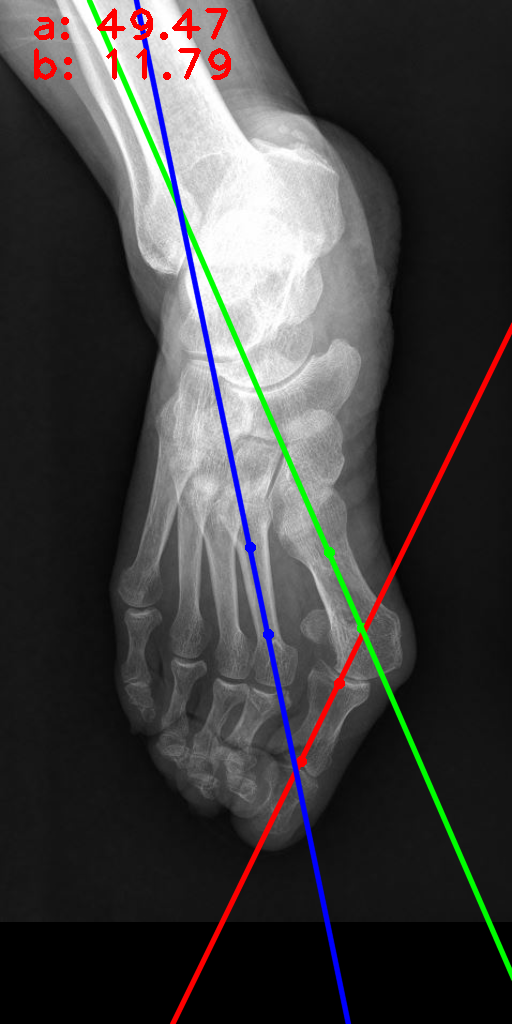}
}
\subfigure[Result]{
    \includegraphics[width=0.1\textwidth]{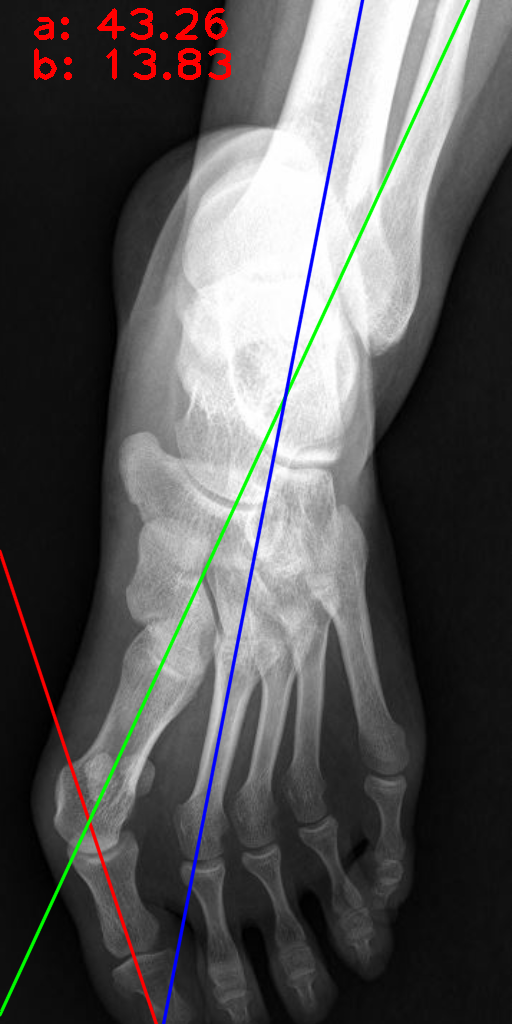}
}
\subfigure[GT]{
    \includegraphics[width=0.1\textwidth]{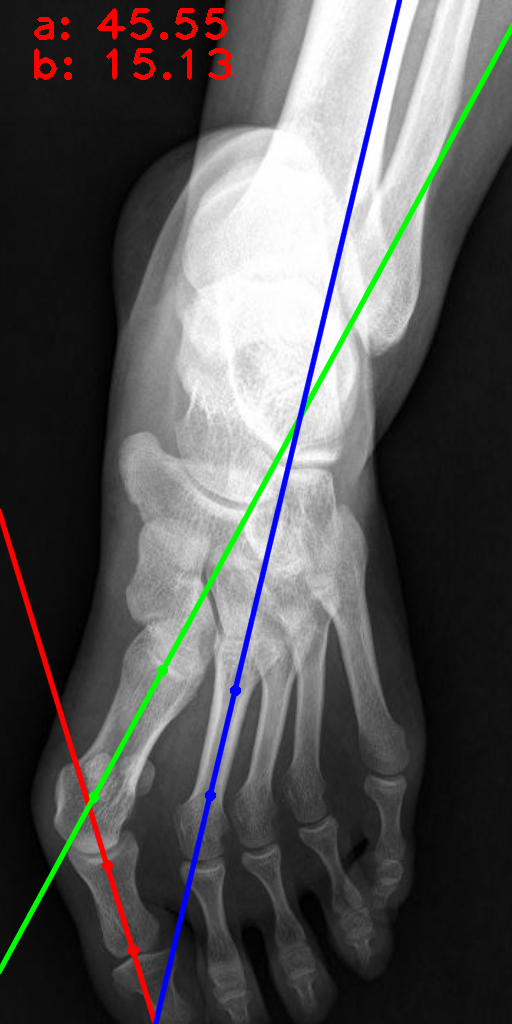}
}
\caption{Results with $Error_{\alpha} < 3^\circ$}
\label{fig:4.2}
\end{figure}

\begin{figure}
\centering
\subfigure[Result]{
    \includegraphics[width=0.1\textwidth]{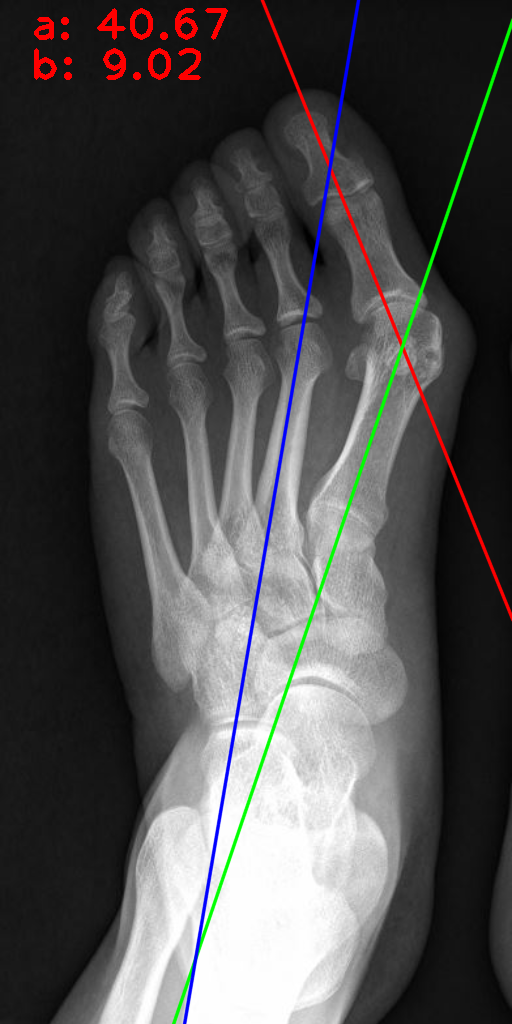}
}
\subfigure[GT]{
    \includegraphics[width=0.1\textwidth]{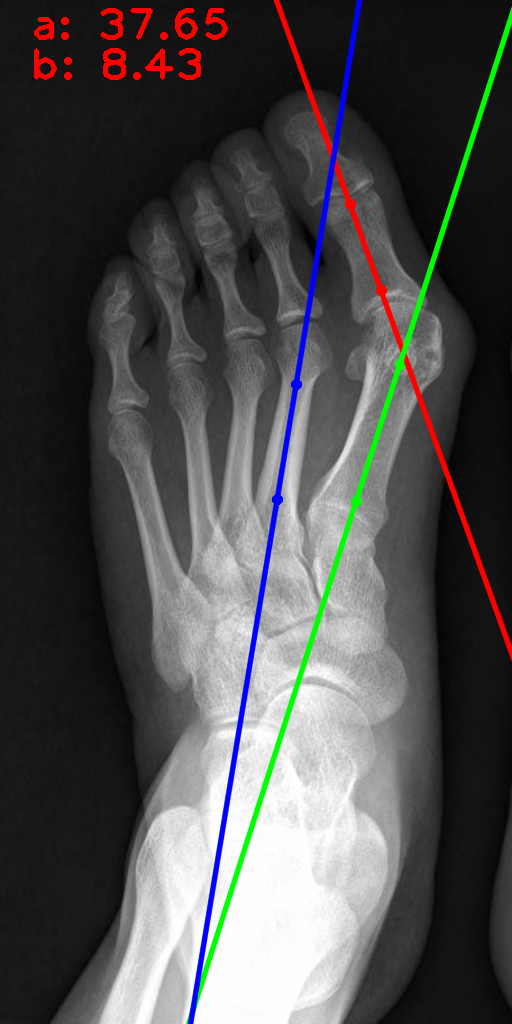}
}
\subfigure[Result]{
    \includegraphics[width=0.1\textwidth]{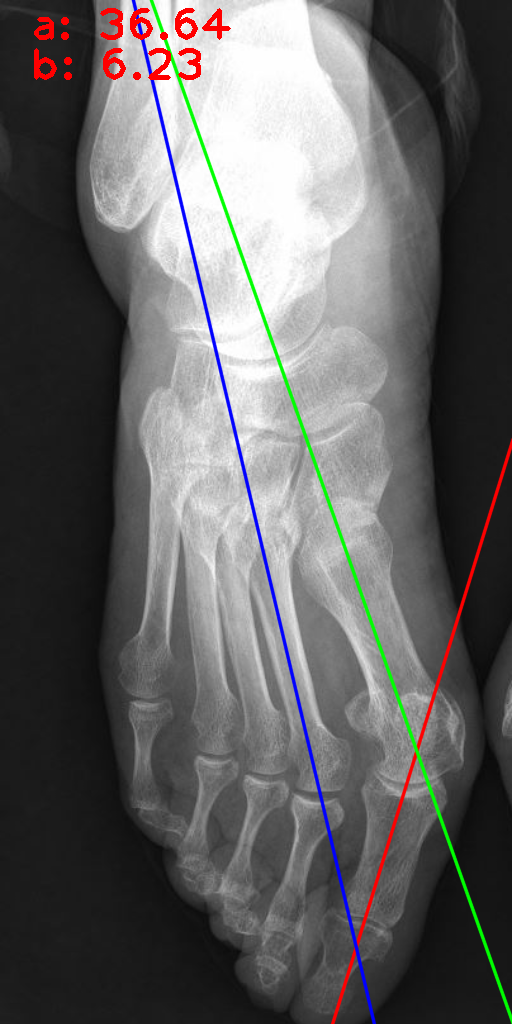}
}
\subfigure[GT]{
    \includegraphics[width=0.1\textwidth]{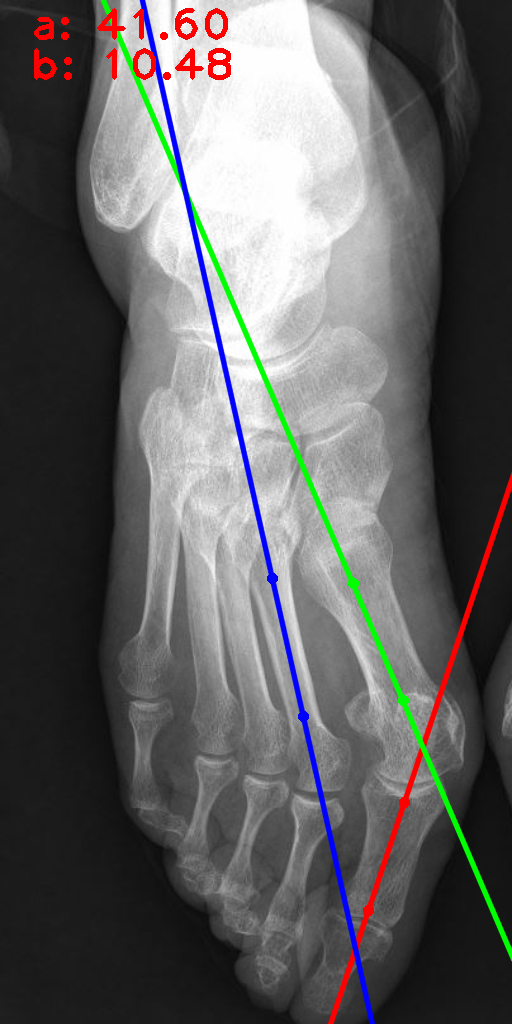}
}
\caption{Results with $3^\circ < Error_{\alpha} < 5^\circ$}
\label{fig:4.3}
\end{figure}

\begin{figure}
\centering
\subfigure[Raw Prediction]{
    \includegraphics[width=0.1\textwidth]{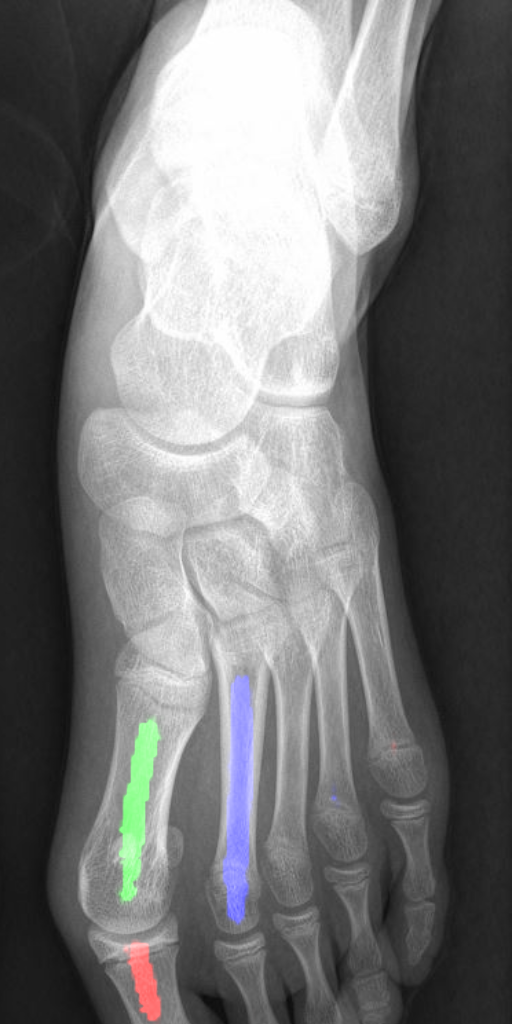}
}
\subfigure[Result]{
    \includegraphics[width=0.1\textwidth]{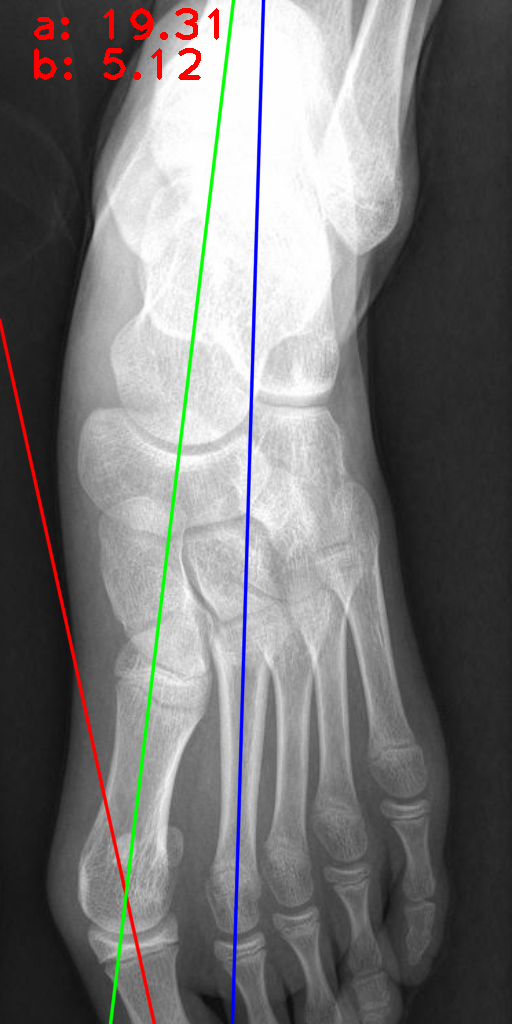}
}
\subfigure[GT]{
    \includegraphics[width=0.1\textwidth]{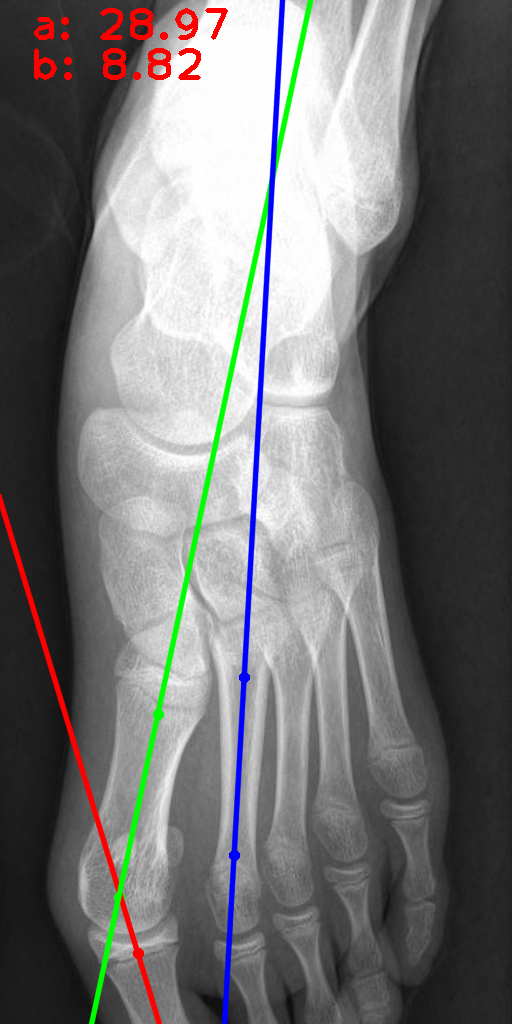}
}

\subfigure[Raw Prediction]{
    \includegraphics[width=0.1\textwidth]{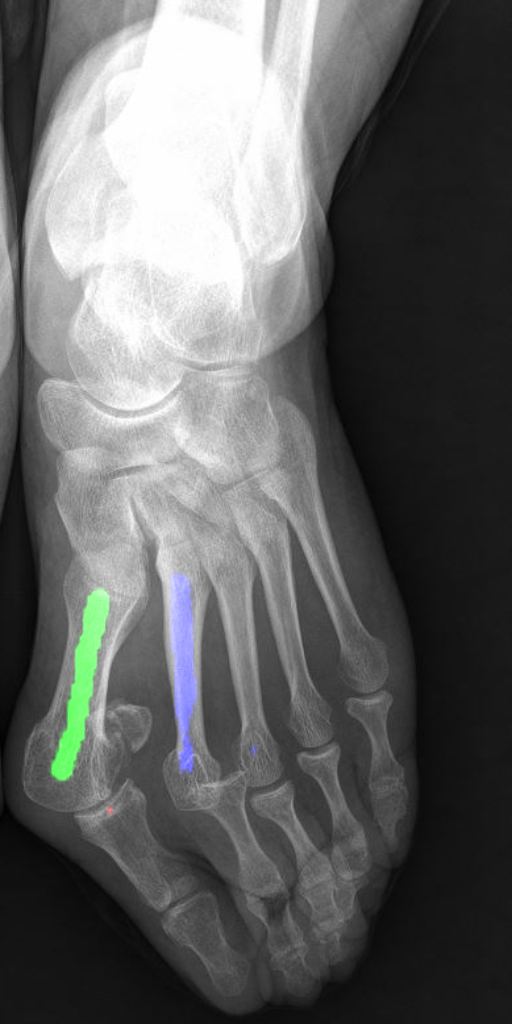}
}
\subfigure[Result]{
    \includegraphics[width=0.1\textwidth]{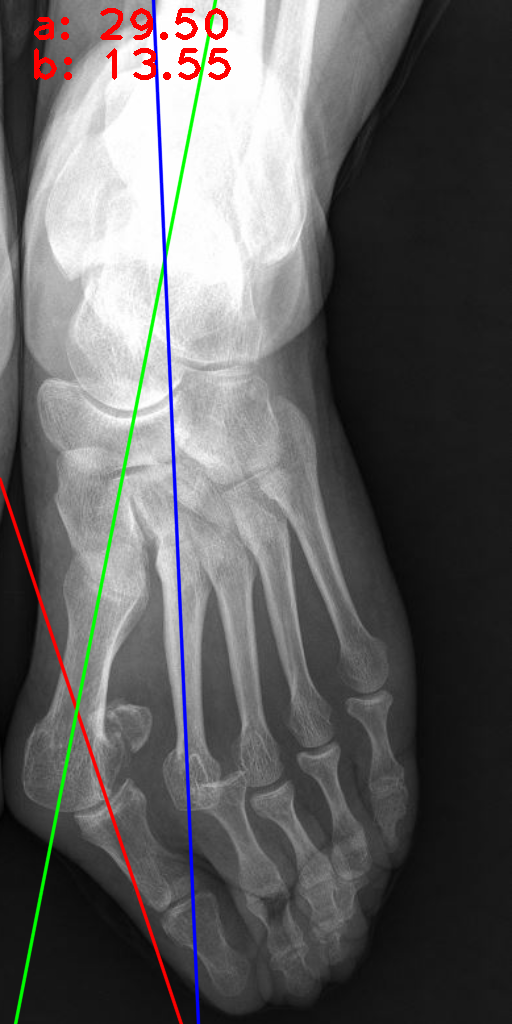}
}
\subfigure[GT]{
    \includegraphics[width=0.1\textwidth]{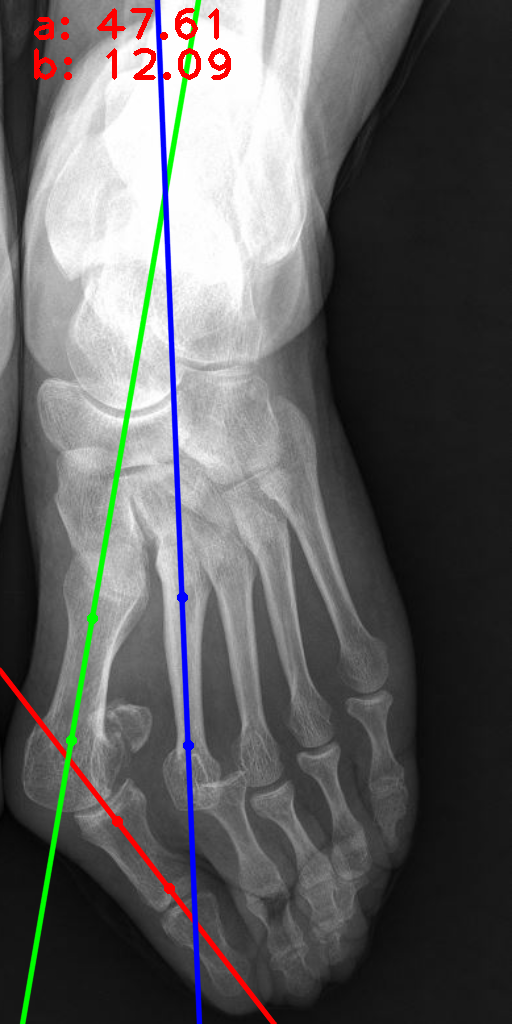}
}
\caption{Results with $Error_{\alpha} > 5^\circ$}
\label{fig:4.4}
\end{figure}

Good results are shown in Fig \ref{fig:4.2}, slightly worse results are shown in Fig \ref{fig:4.3}. Failure cases are shown in \ref{fig:4.4}, it failed because the red ones didn't predicted well. If there are more training samples available, this kind of failure cases will be improved.
\subsection{Impletation Details of our method}
For our neural network, the input size is 1024*512*3 and the output size is 256*128*3. It's worth noting that the original size and aspect ratio of the input images is various.
In order to zoom the image to the same size without distortion, we first scale the image proportionally to a width of 512, then or images with a height greater than 1024, we crop the excess, and for images with a height less than 1024, we fill in black pixels to a height of 1024. The output channels of neural network correspond to three line segments as Fig \ref{fig:3.5} shows.
The optimizer for neural network is RMS, the initial learning rate is 5E-5, the loss function is binary cross entropy, and the activation function of the output layer is sigmoid.
\section{Summary}
To make the HVA measurement more automatic and intelligent, we collect a dataset which include 235 preoperative images from 143 patients.  And we designs a two-stage method combined with deep learning and linear regression, which can carry out high-precision prediction of HVA and IMA.

Future work:
\begin{enumerate}
    \item Making a large-scale foot bone dataset based on virtual simulation.
    \item Based on the transfer learning method, making the large-scale training model in the virtual scene can be well transferred to the real scene.
    \item Extending the scene to postoperative HVA measurements and other bone angle measurements based on X-ray images.
\end{enumerate}

\section{Acknowledgement}
This work was supported by the Ningbo Science and Technology Innovation Project [grant number No.2020Z019]; the Huamei Fund [grant number 2019HMKY10]; and the Zhejiang Medical Fund [grant number 2020KY843].

\bibliographystyle{model1-num-names}

\bibliography{doctor-refs,keyDet-refs}

\begin{thebibliography}{31}
\expandafter\ifx\csname natexlab\endcsname\relax\def\natexlab#1{#1}\fi
\providecommand{\url}[1]{\texttt{#1}}
\providecommand{\href}[2]{#2}
\providecommand{\path}[1]{#1}
\providecommand{\DOIprefix}{doi:}
\providecommand{\ArXivprefix}{arXiv:}
\providecommand{\URLprefix}{URL: }
\providecommand{\Pubmedprefix}{pmid:}
\providecommand{\doi}[1]{\href{http://dx.doi.org/#1}{\path{#1}}}
\providecommand{\Pubmed}[1]{\href{pmid:#1}{\path{#1}}}
\providecommand{\bibinfo}[2]{#2}
\ifx\xfnm\relax \def\xfnm[#1]{\unskip,\space#1}\fi
\bibitem[{Alvarez et~al.(1984)Alvarez, Haddad, Gould, and
  Trevino}]{AlvarezR1984}
\bibinfo{author}{R.~Alvarez}, \bibinfo{author}{R.~J. Haddad},
  \bibinfo{author}{N.~Gould}, \bibinfo{author}{S.~Trevino},
\newblock \bibinfo{title}{The simple bunion: anatomy at the metatarsophalangeal
  joint of the great toe},
\newblock \bibinfo{journal}{Foot \& ankle} \bibinfo{volume}{4}
  (\bibinfo{year}{1984}) \bibinfo{pages}{229--240}.
\bibitem[{Nix et~al.(2010)Nix, Smith, and Vicenzino}]{NixS2010}
\bibinfo{author}{S.~Nix}, \bibinfo{author}{M.~Smith},
  \bibinfo{author}{B.~Vicenzino},
\newblock \bibinfo{title}{Prevalence of hallux valgus in the general
  population: a systematic review and meta-analysis},
\newblock \bibinfo{journal}{Journal of foot and ankle research}
  \bibinfo{volume}{3} (\bibinfo{year}{2010}) \bibinfo{pages}{1--9}.
\bibitem[{Spahn et~al.(2004)Spahn, Schiele, Hell, and Klinger}]{SpahnG2004}
\bibinfo{author}{G.~Spahn}, \bibinfo{author}{R.~Schiele},
  \bibinfo{author}{A.~Hell}, \bibinfo{author}{H.~Klinger},
\newblock \bibinfo{title}{The prevalence of pain and deformities in the feet of
  adolescents. results of a cross-sectional study},
\newblock \bibinfo{journal}{Zeitschrift fur Orthopadie und ihre Grenzgebiete}
  \bibinfo{volume}{142} (\bibinfo{year}{2004}) \bibinfo{pages}{389--396}.
\bibitem[{Benvenuti et~al.(1995)Benvenuti, Ferrucci, Guralnik, Gangemi, and
  Baroni}]{BenvenutiF1995}
\bibinfo{author}{F.~Benvenuti}, \bibinfo{author}{L.~Ferrucci},
  \bibinfo{author}{J.~M. Guralnik}, \bibinfo{author}{S.~Gangemi},
  \bibinfo{author}{A.~Baroni},
\newblock \bibinfo{title}{Foot pain and disability in older persons: an
  epidemiologic survey},
\newblock \bibinfo{journal}{Journal of the American Geriatrics Society}
  \bibinfo{volume}{43} (\bibinfo{year}{1995}) \bibinfo{pages}{479--484}.
\bibitem[{Menz and Lord(2005)}]{MenzHB2005}
\bibinfo{author}{H.~B. Menz}, \bibinfo{author}{S.~R. Lord},
\newblock \bibinfo{title}{Gait instability in older people with hallux valgus},
\newblock \bibinfo{journal}{Foot \& ankle international} \bibinfo{volume}{26}
  (\bibinfo{year}{2005}) \bibinfo{pages}{483--489}.
\bibitem[{Menz and Lord(2001)}]{MenzHB2001}
\bibinfo{author}{H.~B. Menz}, \bibinfo{author}{S.~R. Lord},
\newblock \bibinfo{title}{The contribution of foot problems to mobility
  impairment and falls in community-dwelling older people},
\newblock \bibinfo{journal}{Journal of the American Geriatrics Society}
  \bibinfo{volume}{49} (\bibinfo{year}{2001}) \bibinfo{pages}{1651--1656}.
\bibitem[{Koski et~al.(1996)Koski, Luukinen, Laippala, and Kivela}]{KoskiK1996}
\bibinfo{author}{K.~Koski}, \bibinfo{author}{H.~Luukinen},
  \bibinfo{author}{P.~Laippala}, \bibinfo{author}{S.-L. Kivela},
\newblock \bibinfo{title}{Physiological factors and medications as predictors
  of injurious falls by elderly people: a prospective population-based study},
\newblock \bibinfo{journal}{Age and ageing} \bibinfo{volume}{25}
  (\bibinfo{year}{1996}) \bibinfo{pages}{29--38}.
\bibitem[{Wagner et~al.(2016)Wagner, Ortiz, Torres, Contesse, Vela, and
  Zanolli}]{WagnerE2016}
\bibinfo{author}{E.~Wagner}, \bibinfo{author}{C.~Ortiz},
  \bibinfo{author}{K.~Torres}, \bibinfo{author}{I.~Contesse},
  \bibinfo{author}{O.~Vela}, \bibinfo{author}{D.~Zanolli},
\newblock \bibinfo{title}{Cost effectiveness of different techniques in hallux
  valgus surgery},
\newblock \bibinfo{journal}{Foot and Ankle Surgery} \bibinfo{volume}{22}
  (\bibinfo{year}{2016}) \bibinfo{pages}{259--264}.
\bibitem[{Heineman et~al.(2020)Heineman, Liu, Pacicco, Dessouky, Wukich, and
  Chhabra}]{HeinemanN2020}
\bibinfo{author}{N.~Heineman}, \bibinfo{author}{G.~Liu},
  \bibinfo{author}{T.~Pacicco}, \bibinfo{author}{R.~Dessouky},
  \bibinfo{author}{D.~K. Wukich}, \bibinfo{author}{A.~Chhabra},
\newblock \bibinfo{title}{Clinical and imaging assessment and treatment of
  hallux valgus},
\newblock \bibinfo{journal}{Acta Radiologica} \bibinfo{volume}{61}
  (\bibinfo{year}{2020}) \bibinfo{pages}{56--66}.
\bibitem[{Piqu{\'e}-Vidal and Vila(2009)}]{Pique-VidalCarlos2019}
\bibinfo{author}{C.~Piqu{\'e}-Vidal}, \bibinfo{author}{J.~Vila},
\newblock \bibinfo{title}{A geometric analysis of hallux valgus: correlation
  with clinical assessment of severity},
\newblock \bibinfo{journal}{Journal of foot and ankle research}
  \bibinfo{volume}{2} (\bibinfo{year}{2009}) \bibinfo{pages}{1--8}.
\bibitem[{Lee et~al.(2012)Lee, Ahn, Chung, Sung, and Park}]{LeeKM2012}
\bibinfo{author}{K.~M. Lee}, \bibinfo{author}{S.~Ahn}, \bibinfo{author}{C.~Y.
  Chung}, \bibinfo{author}{K.~H. Sung}, \bibinfo{author}{M.~S. Park},
\newblock \bibinfo{title}{Reliability and relationship of radiographic
  measurements in hallux valgus},
\newblock \bibinfo{journal}{Clinical Orthopaedics and Related
  Research{\textregistered}} \bibinfo{volume}{470} (\bibinfo{year}{2012})
  \bibinfo{pages}{2613--2621}.
\bibitem[{Coughlin and Freund(2001)}]{CoughlinMJ2001}
\bibinfo{author}{M.~J. Coughlin}, \bibinfo{author}{E.~Freund},
\newblock \bibinfo{title}{The reliability of angular measurements in hallux
  valgus deformities},
\newblock \bibinfo{journal}{Foot \& ankle international} \bibinfo{volume}{22}
  (\bibinfo{year}{2001}) \bibinfo{pages}{369--379}.
\bibitem[{Chi et~al.(2002)Chi, Davitt, Younger, Holt, and
  Sangeorzan}]{ChiTD2002}
\bibinfo{author}{T.~D. Chi}, \bibinfo{author}{J.~Davitt},
  \bibinfo{author}{A.~Younger}, \bibinfo{author}{S.~Holt},
  \bibinfo{author}{B.~J. Sangeorzan},
\newblock \bibinfo{title}{Intra-and inter-observer reliability of the distal
  metatarsal articular angle in adult hallux valgus},
\newblock \bibinfo{journal}{Foot \& ankle international} \bibinfo{volume}{23}
  (\bibinfo{year}{2002}) \bibinfo{pages}{722--726}.
\bibitem[{Cruz et~al.(2017)Cruz, Wagner, Henning, Sanhudo, Pagnussato, and
  Galia}]{CruzEP2017}
\bibinfo{author}{E.~P. Cruz}, \bibinfo{author}{F.~V. Wagner},
  \bibinfo{author}{C.~Henning}, \bibinfo{author}{J.~A.~V. Sanhudo},
  \bibinfo{author}{F.~Pagnussato}, \bibinfo{author}{C.~R. Galia},
\newblock \bibinfo{title}{Comparison between simple radiographic and computed
  tomographic three-dimensional reconstruction for evaluation of the distal
  metatarsal articular angle},
\newblock \bibinfo{journal}{The Journal of Foot and Ankle Surgery}
  \bibinfo{volume}{56} (\bibinfo{year}{2017}) \bibinfo{pages}{505--509}.
\bibitem[{van~der Woude et~al.(2019)van~der Woude, Keizer, Wever-Korevaar, and
  Thomassen}]{vanDerWoudeP2019}
\bibinfo{author}{P.~van~der Woude}, \bibinfo{author}{S.~B. Keizer},
  \bibinfo{author}{M.~Wever-Korevaar}, \bibinfo{author}{B.~J. Thomassen},
\newblock \bibinfo{title}{Intra-and interobserver agreement in hallux valgus
  angle measurements on weightbearing and non-weightbearing radiographs},
\newblock \bibinfo{journal}{The Journal of Foot and Ankle Surgery}
  \bibinfo{volume}{58} (\bibinfo{year}{2019}) \bibinfo{pages}{706--712}.
\bibitem[{Dantone et~al.(2013)Dantone, Gall, Leistner, and
  Van~Gool}]{DantoneM2013}
\bibinfo{author}{M.~Dantone}, \bibinfo{author}{J.~Gall},
  \bibinfo{author}{C.~Leistner}, \bibinfo{author}{L.~Van~Gool},
\newblock \bibinfo{title}{Human pose estimation using body parts dependent
  joint regressors},
\newblock in: \bibinfo{booktitle}{Proceedings of the IEEE Conference on
  Computer Vision and Pattern Recognition}, \bibinfo{year}{2013}, pp.
  \bibinfo{pages}{3041--3048}.
\bibitem[{Gkioxari et~al.(2013)Gkioxari, Arbel{\'a}ez, Bourdev, and
  Malik}]{GkioxariG2013}
\bibinfo{author}{G.~Gkioxari}, \bibinfo{author}{P.~Arbel{\'a}ez},
  \bibinfo{author}{L.~Bourdev}, \bibinfo{author}{J.~Malik},
\newblock \bibinfo{title}{Articulated pose estimation using discriminative
  armlet classifiers},
\newblock in: \bibinfo{booktitle}{Proceedings of the IEEE Conference on
  Computer Vision and Pattern Recognition}, \bibinfo{year}{2013}, pp.
  \bibinfo{pages}{3342--3349}.
\bibitem[{Toshev and Szegedy(2014)}]{ToshevA2014}
\bibinfo{author}{A.~Toshev}, \bibinfo{author}{C.~Szegedy},
\newblock \bibinfo{title}{Deeppose: Human pose estimation via deep neural
  networks},
\newblock in: \bibinfo{booktitle}{Proceedings of the IEEE conference on
  computer vision and pattern recognition}, \bibinfo{year}{2014}, pp.
  \bibinfo{pages}{1653--1660}.
\bibitem[{Carreira et~al.(2016)Carreira, Agrawal, Fragkiadaki, and
  Malik}]{CarreiraJ2016}
\bibinfo{author}{J.~Carreira}, \bibinfo{author}{P.~Agrawal},
  \bibinfo{author}{K.~Fragkiadaki}, \bibinfo{author}{J.~Malik},
\newblock \bibinfo{title}{Human pose estimation with iterative error feedback},
\newblock in: \bibinfo{booktitle}{Proceedings of the IEEE conference on
  computer vision and pattern recognition}, \bibinfo{year}{2016}, pp.
  \bibinfo{pages}{4733--4742}.
\bibitem[{Sun et~al.(2017)Sun, Shang, Liang, and Wei}]{SunX2017}
\bibinfo{author}{X.~Sun}, \bibinfo{author}{J.~Shang},
  \bibinfo{author}{S.~Liang}, \bibinfo{author}{Y.~Wei},
\newblock \bibinfo{title}{Compositional human pose regression},
\newblock in: \bibinfo{booktitle}{Proceedings of the IEEE International
  Conference on Computer Vision}, \bibinfo{year}{2017}, pp.
  \bibinfo{pages}{2602--2611}.
\bibitem[{Newell et~al.(2016)Newell, Yang, and Deng}]{NewellA2016}
\bibinfo{author}{A.~Newell}, \bibinfo{author}{K.~Yang},
  \bibinfo{author}{J.~Deng},
\newblock \bibinfo{title}{Stacked hourglass networks for human pose
  estimation},
\newblock in: \bibinfo{booktitle}{European conference on computer vision},
  \bibinfo{organization}{Springer}, \bibinfo{year}{2016}, pp.
  \bibinfo{pages}{483--499}.
\bibitem[{Wei et~al.(2016)Wei, Ramakrishna, Kanade, and Sheikh}]{WeiSE2016}
\bibinfo{author}{S.-E. Wei}, \bibinfo{author}{V.~Ramakrishna},
  \bibinfo{author}{T.~Kanade}, \bibinfo{author}{Y.~Sheikh},
\newblock \bibinfo{title}{Convolutional pose machines},
\newblock in: \bibinfo{booktitle}{Proceedings of the IEEE conference on
  Computer Vision and Pattern Recognition}, \bibinfo{year}{2016}, pp.
  \bibinfo{pages}{4724--4732}.
\bibitem[{Tompson et~al.(2014)Tompson, Jain, LeCun, and Bregler}]{TompsonJ2014}
\bibinfo{author}{J.~Tompson}, \bibinfo{author}{A.~Jain},
  \bibinfo{author}{Y.~LeCun}, \bibinfo{author}{C.~Bregler},
\newblock \bibinfo{title}{Joint training of a convolutional network and a
  graphical model for human pose estimation},
\newblock \bibinfo{journal}{arXiv preprint arXiv:1406.2984}
  (\bibinfo{year}{2014}).
\bibitem[{Papandreou et~al.(2017)Papandreou, Zhu, Kanazawa, Toshev, Tompson,
  Bregler, and Murphy}]{PapandreouG2017}
\bibinfo{author}{G.~Papandreou}, \bibinfo{author}{T.~Zhu},
  \bibinfo{author}{N.~Kanazawa}, \bibinfo{author}{A.~Toshev},
  \bibinfo{author}{J.~Tompson}, \bibinfo{author}{C.~Bregler},
  \bibinfo{author}{K.~Murphy},
\newblock \bibinfo{title}{Towards accurate multi-person pose estimation in the
  wild},
\newblock in: \bibinfo{booktitle}{Proceedings of the IEEE Conference on
  Computer Vision and Pattern Recognition}, \bibinfo{year}{2017}, pp.
  \bibinfo{pages}{4903--4911}.
\bibitem[{Rafi et~al.(2016)Rafi, Leibe, Gall, and Kostrikov}]{RafiU2016}
\bibinfo{author}{U.~Rafi}, \bibinfo{author}{B.~Leibe},
  \bibinfo{author}{J.~Gall}, \bibinfo{author}{I.~Kostrikov},
\newblock \bibinfo{title}{An efficient convolutional network for human pose
  estimation.},
\newblock in: \bibinfo{booktitle}{BMVC}, volume~\bibinfo{volume}{1},
  \bibinfo{year}{2016}, p.~\bibinfo{pages}{2}.
\bibitem[{Xiao et~al.(2018)Xiao, Wu, and Wei}]{XiaoB2018}
\bibinfo{author}{B.~Xiao}, \bibinfo{author}{H.~Wu}, \bibinfo{author}{Y.~Wei},
\newblock \bibinfo{title}{Simple baselines for human pose estimation and
  tracking},
\newblock in: \bibinfo{booktitle}{Proceedings of the European conference on
  computer vision (ECCV)}, \bibinfo{year}{2018}, pp. \bibinfo{pages}{466--481}.
\bibitem[{Tompson et~al.(2015)Tompson, Goroshin, Jain, LeCun, and
  Bregler}]{TompsonJ2015}
\bibinfo{author}{J.~Tompson}, \bibinfo{author}{R.~Goroshin},
  \bibinfo{author}{A.~Jain}, \bibinfo{author}{Y.~LeCun},
  \bibinfo{author}{C.~Bregler},
\newblock \bibinfo{title}{Efficient object localization using convolutional
  networks},
\newblock in: \bibinfo{booktitle}{Proceedings of the IEEE conference on
  computer vision and pattern recognition}, \bibinfo{year}{2015}, pp.
  \bibinfo{pages}{648--656}.
\bibitem[{Bulat and Tzimiropoulos(2016)}]{BulatA2016}
\bibinfo{author}{A.~Bulat}, \bibinfo{author}{G.~Tzimiropoulos},
\newblock \bibinfo{title}{Human pose estimation via convolutional part heatmap
  regression},
\newblock in: \bibinfo{booktitle}{European Conference on Computer Vision},
  \bibinfo{organization}{Springer}, \bibinfo{year}{2016}, pp.
  \bibinfo{pages}{717--732}.
\bibitem[{Yang et~al.(2017)Yang, Li, Ouyang, Li, and Wang}]{YangW2017}
\bibinfo{author}{W.~Yang}, \bibinfo{author}{S.~Li},
  \bibinfo{author}{W.~Ouyang}, \bibinfo{author}{H.~Li},
  \bibinfo{author}{X.~Wang},
\newblock \bibinfo{title}{Learning feature pyramids for human pose estimation},
\newblock in: \bibinfo{booktitle}{proceedings of the IEEE international
  conference on computer vision}, \bibinfo{year}{2017}, pp.
  \bibinfo{pages}{1281--1290}.
\bibitem[{Belagiannis and Zisserman(2017)}]{BelagiannisV2017}
\bibinfo{author}{V.~Belagiannis}, \bibinfo{author}{A.~Zisserman},
\newblock \bibinfo{title}{Recurrent human pose estimation},
\newblock in: \bibinfo{booktitle}{2017 12th IEEE International Conference on
  Automatic Face \& Gesture Recognition (FG 2017)},
  \bibinfo{organization}{IEEE}, \bibinfo{year}{2017}, pp.
  \bibinfo{pages}{468--475}.
\bibitem[{Sun et~al.(2019)Sun, Xiao, Liu, and Wang}]{SunK2019}
\bibinfo{author}{K.~Sun}, \bibinfo{author}{B.~Xiao}, \bibinfo{author}{D.~Liu},
  \bibinfo{author}{J.~Wang},
\newblock \bibinfo{title}{Deep high-resolution representation learning for
  human pose estimation},
\newblock in: \bibinfo{booktitle}{Proceedings of the IEEE/CVF Conference on
  Computer Vision and Pattern Recognition}, \bibinfo{year}{2019}, pp.
  \bibinfo{pages}{5693--5703}.

\end{thebibliography}


\end{document}